\documentclass{article}
\usepackage{arxiv}

\usepackage[utf8]{inputenc} 
\usepackage[T1]{fontenc}    
\usepackage[hidelinks]{hyperref}       
\usepackage{hyphenat}
\usepackage{amsmath}
\usepackage{amssymb}
\usepackage{url}            
\usepackage{booktabs}       
\usepackage{tabularx}       
\usepackage{amsfonts}       
\usepackage{nicefrac}       
\usepackage{microtype}      
\usepackage{graphicx}
\usepackage{doi}

\usepackage{longtable}
\usepackage{array}
\usepackage{booktabs}

\usepackage{tikz}
\usepackage{pgfplots}
\pgfplotsset{compat=1.18}

\usetikzlibrary{arrows.meta}
\usetikzlibrary{positioning}
\usetikzlibrary{shapes.geometric}
\usetikzlibrary{trees}
\usetikzlibrary{calc}

\title{KV Cache Optimization Strategies for Scalable\\and Efficient LLM Inference}

\author{\href{https://orcid.org/0009-0003-8849-610X}{Yichun~Xu$^1$ \includegraphics[scale=0.07]{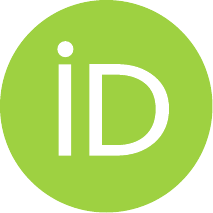}},~\href{https://orcid.org/0000-0002-0035-1068}{Navjot~K.~Khaira$^2$ \includegraphics[scale=0.07]{orcid.pdf}}\normalfont{,~and}~\bfseries \href{https://orcid.org/0000-0002-5870-6204}{Tejinder~Singh$^2$ \includegraphics[scale=0.07]{orcid.pdf}}\vspace{2mm} \\ 
	$^1$Dell Technologies, Hopkinton, MA 01748, USA \\
    $^2$Dell Technologies, Santa Clara, CA 95054, USA\vspace{2mm} \\
	\texttt{\{Yichun.Xu, Navjot.Khaira, Singh.Tejinder\}@Dell.com} \\
}

\newcolumntype{L}{>{\raggedright\arraybackslash}X}

\renewcommand{\headeright}{}
\renewcommand{\undertitle}{}
\renewcommand{\shorttitle}{\textbf{Y. Xu}, \textbf{N. K. Khaira}, and \textbf{T. Singh}: KV Cache Optimization Strategies for Scalable and Efficient LLM Inference}

\hypersetup{
pdftitle={KV Cache Optimization Strategies for Scalable and Efficient LLM Inference},
pdfsubject={Computer Science, Artificial Intelligence},
pdfauthor={Yichun~Xu, Navjot K.~Khaira, Tejinder~Singh},
pdfkeywords={Key-Value Cache Management, LLMs Efficiency, Transformer Memory Optimization, Attention Mechanism Compression, Memory Bandwidth Optimization},
}

\begin{document}
\maketitle

\begin{abstract}
The key-value (KV) cache is a foundational optimization in Transformer-based large language models (LLMs), eliminating redundant recomputation of past token representations during autoregressive generation. However, its memory footprint scales linearly with context length, imposing critical bottlenecks on GPU memory capacity, memory bandwidth, and inference throughput as production LLMs push context windows from thousands to millions of tokens. Efficient KV cache management has thus become a first-order challenge for scalable LLM deployment. This paper provides a systematic review of recent KV cache optimization techniques, organizing them into five principal directions: cache eviction, cache compression, hybrid memory solutions, novel attention mechanisms, and combination strategies. For each category we analyze the underlying mechanisms, deployment trade-offs, and empirical performance across memory reduction, throughput, and model accuracy metrics. We further map techniques to seven practical deployment scenarios, including long-context single requests, high-throughput datacenter serving, edge devices, multi-turn conversations, and accuracy-critical reasoning, providing actionable guidance for practitioners selecting among competing approaches. Our analysis reveals that no single technique dominates across all settings; instead, the optimal strategy depends on context length, hardware constraints, and workload characteristics, pointing toward adaptive, multi-stage optimization pipelines as a promising direction for future research.
\end{abstract}


\keywords{Key-Value Cache Management\and
LLMs Efficiency \and
Transformer Memory Optimization \and
Attention Mechanism Compression \and 
Memory Bandwidth Optimization}

\section{Introduction}
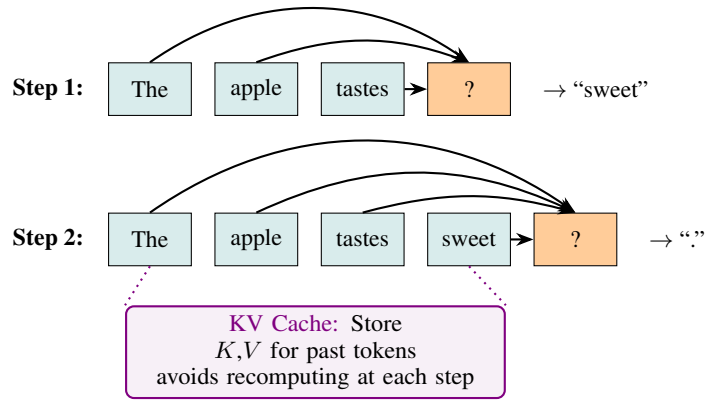
\begin{figure}[htbp]
\centering
\begin{tikzpicture}[
  token/.style={rectangle, draw, fill=teal!15,
                minimum width=1.1cm, minimum height=0.7cm,
                font=\small},
  newtoken/.style={rectangle, draw, fill=orange!40,
                   minimum width=1.1cm, minimum height=0.7cm,
                   font=\small},
  arrow/.style={-{Stealth}, thick},
  node distance=0.3cm
]
\node[token] (t1a) at (0, 2) {The};
\node[token, right=of t1a] (t2a) {apple};
\node[token, right=of t2a] (t3a) {tastes};
\node[newtoken, right=of t3a] (t4a) {?};
\node[left=0.2cm of t1a, font=\small\bfseries] {Step 1:};
\draw[arrow, bend left=35] (t1a.north) to (t4a.north);
\draw[arrow, bend left=20] (t2a.north) to (t4a.north);
\draw[arrow] (t3a) -- (t4a);
\node[right=0.3cm of t4a, font=\small] {$\rightarrow$ ``sweet''};
\node[token] (t1b) at (0, 0) {The};
\node[token, right=of t1b] (t2b) {apple};
\node[token, right=of t2b] (t3b) {tastes};
\node[token, right=of t3b] (t4b) {sweet};
\node[newtoken, right=of t4b] (t5b) {?};
\node[left=0.2cm of t1b, font=\small\bfseries] {Step 2:};
\draw[arrow, bend left=35] (t1b.north) to (t5b.north);
\draw[arrow, bend left=25] (t2b.north) to (t5b.north);
\draw[arrow, bend left=15] (t3b.north) to (t5b.north);
\draw[arrow] (t4b) -- (t5b);
\node[right=0.3cm of t5b, font=\small] {$\rightarrow$ ``.''};
\node[draw=violet, thick, rounded corners=3pt, fill=violet!6,
      font=\small, align=center, text width=4.8cm]
  (ann) at (2.2, -1.5)
  {\textcolor{violet}{KV Cache:} Store $K$,$V$ for past tokens\\
   avoids recomputing at each step};
\draw[violet, dotted, thick] (t1b.south) -- (ann.north west);
\draw[violet, dotted, thick] (t4b.south) -- (ann.north east);
\end{tikzpicture}
\caption{Autoregressive generation, at each step the new token (orange) attends to all prior tokens (cyan). Without caching, keys and values for every past token would be recomputed from scratch at each step. The KV cache avoids this by storing and reusing them.}
\label{fig:autoreg}
\end{figure}

KV cache optimization is crucial for LLM deployment. It reduces latency by avoiding recomputing past tokens as shown in Fig~\ref{fig:autoreg}, lowers operational costs with less compute requirements, and enables deployment on resource-constrained devices by managing large memory footprints and optimizing data transfers. Recent advancements in LLMs illustrate a rapid and industry‑wide increase in supported context window sizes. In recent years, modern models across different vendors have expanded from tens of thousands of tokens to hundreds of thousands, and in some cases millions. This reflects a broader trend rather than the evolution of any single product line, underscoring the growing importance of efficient KV‑cache management in large context inference.

Unlike individual research papers that typically focus on one KV‑cache optimization method in isolation, or prior surveys that provide broad but shallow coverage, this survey offers a middle-ground perspective. Our work  systematically reviews and categorizes recent strategies for KV cache optimization into five major directions: cache eviction, compression and reconstruction, hybrid memory solutions, novel attention mechanisms, and combined approaches.  Each category presents unique trade-offs between memory efficiency, computational cost, and model accuracy. Representative methods such as H$_2$O, SnapKV, and Ada-KV demonstrate the effectiveness of intelligent token selection \cite{H2O, SnapKV, Ada-KV, AIBenchmarking}, while compression techniques like KIVI highlight the potential of quantization \cite{KIVI}. 

This paper focuses on the most recent approaches in each category, providing detailed explanations of their mechanisms rather than simply brief mentions. This approach ensures clarity and depth, enabling readers to understand the principles behind each method and their implications for future research. 

\section{Background} \label{sec:background}

\subsection{Context Length} \label{subsec:context_length}
The concept of a fixed input sequence length (context length) was first introduced in Attention Is All You Need \cite{Attention}. The authors proposed the Transformer architecture and defined the usage of a fixed-size input sequence length due to computational constraints of the self-attention mechanism. The term ``context length'' and ``context window'' subsequently became standard terminology in the machine learning to refer to this fixed limit on the number of tokens a model could process at one time. For instance, a model with a context length of 2,048 tokens can attend to up to 2,048 tokens at once. This parameter is typically documented in model cards and varies across architectures.  Because attention requires storing keys and values for every past token, increasing context length directly inflates KV cache size and bandwidth demands. As a result, larger context lengths impose higher memory consumption and communication overhead, especially in distributed inference settings.

\subsection{KV Cache}
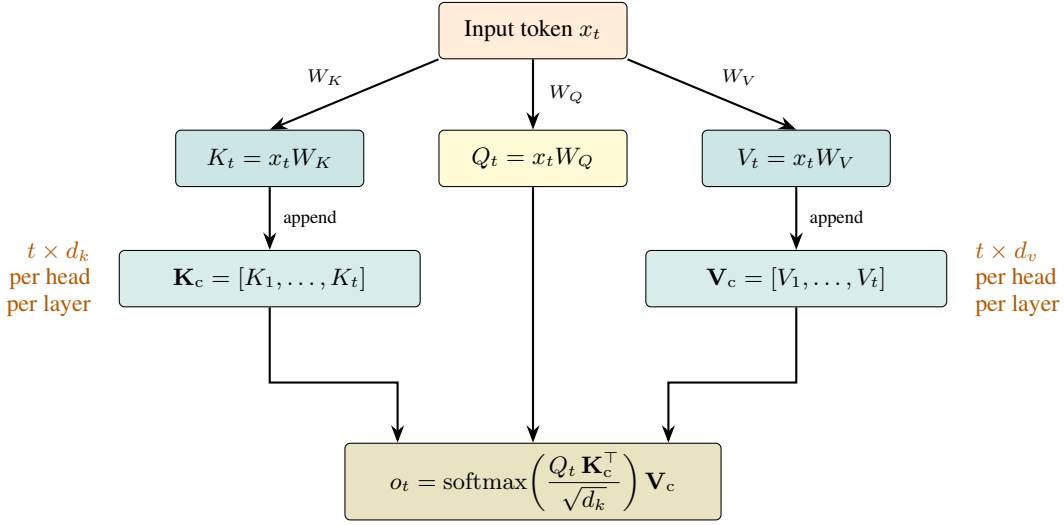
\begin{figure}[htbp]
\centering
\begin{tikzpicture}[
  box/.style={rectangle, draw, rounded corners=2pt,
              minimum width=2.5cm, minimum height=0.75cm,
              font=\small, align=center},
  kv/.style={rectangle, draw, fill=teal!15, rounded corners=2pt,
             minimum width=4.0cm, minimum height=0.75cm,
             font=\small, align=center},
  outbox/.style={rectangle, draw, fill=olive!20, rounded corners=2pt,
              minimum width=5cm, minimum height=0.9cm,
              font=\small, align=center},
  arr/.style={-{Stealth}, thick}
]

\node[box, fill=orange!15] (xt) at (4.5, 3.5) {Input token $x_t$};

\node[box, fill=teal!20]   (K) at (1.0, 1.8) {$K_t = x_t W_K$};
\node[box, fill=yellow!20] (Q) at (4.5, 1.8) {$Q_t = x_t W_Q$};
\node[box, fill=teal!20]   (V) at (8.0, 1.8) {$V_t = x_t W_V$};

\node[kv] (Kc) at (1.0, 0.2) {$\mathbf{K}_\mathrm{c} = [K_1,\ldots,K_t]$};
\node[kv] (Vc) at (8.0, 0.2) {$\mathbf{V}_\mathrm{c} = [V_1,\ldots,V_t]$};

\node[font=\footnotesize, text=orange!70!black, align=right,
      left=0.25cm of Kc]
  {$t \times d_k$\\per head\\per layer};
\node[font=\footnotesize, text=orange!70!black, align=left,
      right=0.25cm of Vc]
  {$t \times d_v$\\per head\\per layer};

\node[outbox] (attn) at (4.5, -2.5)
  {$o_t = \mathrm{softmax}\!\left(\dfrac{Q_t\,\mathbf{K}_\mathrm{c}^{\top}}{\sqrt{d_k}}\right)\mathbf{V}_\mathrm{c}$};

\draw[arr] (xt.south west) -- (K.north)
  node[midway, above left, font=\scriptsize] {$W_K$};
\draw[arr] (xt.south) -- (Q.north)
  node[midway, right=0.08cm, font=\scriptsize] {$W_Q$};
\draw[arr] (xt.south east) -- (V.north)
  node[midway, above right, font=\scriptsize] {$W_V$};

\draw[arr] (K.south) -- (Kc.north)
  node[midway, right=0.05cm, font=\scriptsize] {append};
\draw[arr] (V.south) -- (Vc.north)
  node[midway, right=0.05cm, font=\scriptsize] {append};

\draw[arr] (Q.south) -- (attn.north);

\draw[arr] (Kc.south) -- ++(0, -1.0)
  -| ($(attn.north) + (-1.8, 0)$);

\draw[arr] (Vc.south) -- ++(0, -1.0)
  -| ($(attn.north) + (1.8, 0)$);

\end{tikzpicture}
\caption{Data-flow of the KV cache within a single transformer layer.
  Input token $x_t$ fans into three projections;
  $K_t$ and $V_t$ are appended to their respective caches (teal);
  $Q_t$ attends over the full caches to produce output $o_t$.
  Cache size grows as $O(T)$ per head per layer.}
\label{fig:kvcache_structure}
\end{figure}

The key value (KV) cache is an inherent optimization of the Transformer architecture. It stores intermediate key and value vectors from previous tokens to avoid redundant computation during generation. The key vector represents the identity of a token, while the value vector represents the actual content of that token. As new tokens are generated, their corresponding keys and values are appended to the existing cache. In this way, KV cache can help reduce latency and computational overhead. Fig.~\ref{fig:kvcache_structure} illustrates this
data-flow within a single transformer layer.

While KV caching accelerates inference, its memory footprint grows linearly with context length, since each token in the context is represented by a key value pair. For a Transformer model, the KV cache size can be expressed as:

\begin{equation}
	KV_{per\ token} = 2 \times H\times D \times B \times L
\end{equation}

\qquad  where $H$ is the number of attention heads; $D$ the dimension of each head; $L$ the number of transformer layers; $B$ the bytes per element (typically 4 for \texttt{float32}). Since these parameters are fixed for a given model, the KV per token is constant for a model.

\begin{equation}
    KV_{cache\ size} = KV_{per\ token} \times \mathrm{Context Length}
\end{equation}

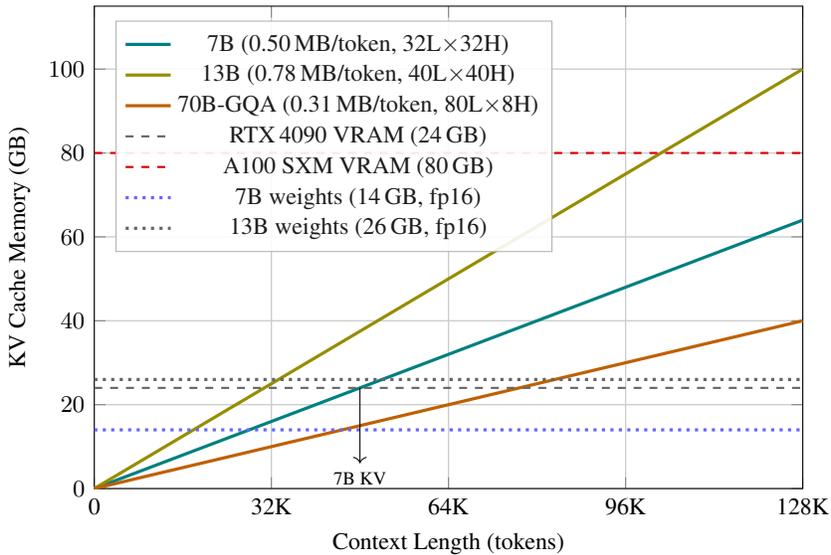
\begin{figure}[htbp]
\centering
\begin{tikzpicture}
\begin{axis}[
    xlabel={Context Length (tokens)},
    ylabel={KV Cache Memory (GB)},
    width=11cm, height=8cm,
    grid=major,
    grid style={line width=0.5pt, draw=gray!40},
    legend pos=north west,
    legend style={font=\footnotesize, fill=white, fill opacity=0.9,
                  draw=gray!60, inner sep=3pt},
    scaled x ticks=false,
    xtick={0,32000,64000,96000,128000},
    xticklabels={0,32K,64K,96K,128K},
    xmin=0, xmax=128000,
    ymin=0, ymax=115,
    tick label style={font=\small},
    label style={font=\small},
    enlargelimits=false,
    clip=true,
]

\addplot[teal, very thick] coordinates {
    (0,0)(32000,16)(64000,32)(96000,48)(128000,64)
};
\addlegendentry{7B (0.50\,MB/token, 32L$\times$32H)};

\addplot[olive, very thick] coordinates {
    (0,0)(32000,25)(64000,50)(96000,75)(128000,100)
};
\addlegendentry{13B (0.78\,MB/token, 40L$\times$40H)};

\addplot[orange!75!black, very thick] coordinates {
    (0,0)(32000,10)(64000,20)(96000,30)(128000,40)
};
\addlegendentry{70B-GQA (0.31\,MB/token, 80L$\times$8H)};

\addplot[gray!80!black, dashed, thick] coordinates {
    (0,24)(128000,24)
};
\addlegendentry{RTX\,4090 VRAM (24\,GB)}

\addplot[red!85!black, dashed, thick] coordinates {
    (0,80)(128000,80)
};
\addlegendentry{A100 SXM VRAM (80\,GB)}

\addplot[blue!60, dotted, very thick] coordinates {
    (0,14)(128000,14)
};
\addlegendentry{7B weights (14\,GB, fp16)}

\addplot[black!60, dotted, very thick] coordinates {
    (0,26)(128000,26)
};
\addlegendentry{13B weights (26\,GB, fp16)}

\draw[black, thin, ->] (48000,24) -- (48000,6)
  node[below, font=\scriptsize, align=center]
  {7B KV\\exceeds\\RTX\,4090\\at $\approx$48K};

\end{axis}
\end{tikzpicture}
\caption{KV cache memory as a function of context length for three LLaMA-2 model variants under fp16 precision. Dashed lines mark GPU VRAM limits; dotted lines mark model parameter memory. At 128K tokens, a 7B model's KV cache ($\approx$64\,GB) exceeds the capacity of an A100 GPU, illustrating the memory bottleneck that motivates KV cache optimization. Values computed as $2 \times L \times H_{\text{kv}} \times d_h \times 2$\,bytes per token; 70B uses GQA with 8 KV heads.}
\label{fig:kv_linear_growth}
\end{figure}

With modern LLMs extending context windows from thousands to millions of tokens, this linear growth imposes significant memory and bandwidth demands,
making KV cache optimization essential for efficient deployment, as quantified in Fig.~\ref{fig:kv_linear_growth} for three representative model sizes.

\subsection{Attention Score}
The Transformer architecture relies on an attention mechanism to determine the relevance of each token in a sequence when generating the next token.  Attention score is calculated to quantify how much a token should attend to other tokens in the input. For example, when generating the word “\texttt{sweet}” in the sentence “\texttt{The apple tastes sweet}”,  the model assigns higher attention to “\texttt{apple}” than to “\texttt{The}”, as it is more contextually relevant. The most widely used approach for computing attention scores is Scaled Dot-Product Attention. 

For a query vector $Q_i$ corresponding to token $i$ and a key vector $K_j$ for token $j$, the similarity is computed as:

\begin{equation}
    \mathrm{Score}\left( Q_i, K_j\right) = \frac{Q_i \cdot K_j^{T}}{\sqrt{d_k}}
\end{equation}

\qquad where $d_k$ is the dimension of the hidden layer, also known as the size of vectors used in the model. Dot product represents how much two vectors point in the same direction. In the above formula, dot product of $Qi$ and $Kj^T$ shows the similarity of $i$'s query is to $j$'s key, reflecting how much token $i$ attend to previous token $j$ during generation.  

These scores are normalized using the softmax function to produce attention weights that sum to 1:

\begin{equation}
\alpha_{ij} = \mathrm{softmax}\!\left( \frac{Q_i \cdot K_j^{T}}{\sqrt{d_k}} \right)
\end{equation}

Finally, the output for token $i$ is computed as a weighted sum of the value vectors $V_j$:

\begin{equation}
\mathrm{output}_i = \sum_{j} \alpha_{ij} V_j
\end{equation}

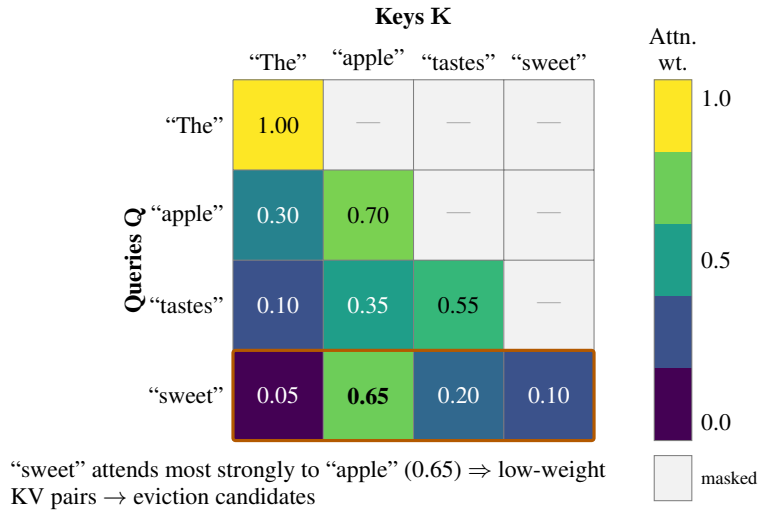
\begin{figure}[htbp]
\centering
\begin{tikzpicture}[font=\small]

\definecolor{virMask}{HTML}{F2F2F2}  
\definecolor{vir005} {HTML}{440154}  
\definecolor{vir010} {HTML}{3B528B}  
\definecolor{vir020} {HTML}{31688E}  
\definecolor{vir030} {HTML}{26828E}  
\definecolor{vir035} {HTML}{1F9E89}  
\definecolor{vir055} {HTML}{35B779}  
\definecolor{vir065} {HTML}{6DCD59}  
\definecolor{vir070} {HTML}{7AD151}  
\definecolor{vir100} {HTML}{FDE725}  


\fill[vir100]  (0,   3.6) rectangle (1.2, 4.8);
\fill[virMask] (1.2, 3.6) rectangle (4.8, 4.8);
\fill[vir030]  (0,   2.4) rectangle (1.2, 3.6);
\fill[vir070]  (1.2, 2.4) rectangle (2.4, 3.6);
\fill[virMask] (2.4, 2.4) rectangle (4.8, 3.6);
\fill[vir010]  (0,   1.2) rectangle (1.2, 2.4);
\fill[vir035]  (1.2, 1.2) rectangle (2.4, 2.4);
\fill[vir055]  (2.4, 1.2) rectangle (3.6, 2.4);
\fill[virMask] (3.6, 1.2) rectangle (4.8, 2.4);
\fill[vir005]  (0,   0)   rectangle (1.2, 1.2);
\fill[vir065]  (1.2, 0)   rectangle (2.4, 1.2);
\fill[vir020]  (2.4, 0)   rectangle (3.6, 1.2);
\fill[vir010]  (3.6, 0)   rectangle (4.8, 1.2);

\draw[black!50, thin] (0,0) grid[xstep=1.2, ystep=1.2] (4.8, 4.8);

\node[text=black]       at (0.6, 4.2) {1.00};
\node[text=black!30]    at (1.8, 4.2) {\textemdash};
\node[text=black!30]    at (3.0, 4.2) {\textemdash};
\node[text=black!30]    at (4.2, 4.2) {\textemdash};
\node[text=white]       at (0.6, 3.0) {0.30};
\node[text=black]       at (1.8, 3.0) {0.70};
\node[text=black!30]    at (3.0, 3.0) {\textemdash};
\node[text=black!30]    at (4.2, 3.0) {\textemdash};
\node[text=white]       at (0.6, 1.8) {0.10};
\node[text=white]       at (1.8, 1.8) {0.35};
\node[text=black]       at (3.0, 1.8) {0.55};
\node[text=black!30]    at (4.2, 1.8) {\textemdash};
\node[text=white]       at (0.6, 0.6) {0.05};
\node[text=black, font=\small\bfseries] at (1.8, 0.6) {0.65};
\node[text=white]       at (3.0, 0.6) {0.20};
\node[text=white]       at (4.2, 0.6) {0.10};

\node[above, font=\footnotesize] at (0.6,  4.8) {``The''};
\node[above, font=\footnotesize] at (1.8,  4.8) {``apple''};
\node[above, font=\footnotesize] at (3.0,  4.8) {``tastes''};
\node[above, font=\footnotesize] at (4.2,  4.8) {``sweet''};
\node[above, font=\small\bfseries] at (2.4, 5.35) {Keys $\mathbf{K}$};

\node[left, font=\footnotesize] at (0, 4.2) {``The''};
\node[left, font=\footnotesize] at (0, 3.0) {``apple''};
\node[left, font=\footnotesize] at (0, 1.8) {``tastes''};
\node[left, font=\footnotesize] at (0, 0.6) {``sweet''};
\node[rotate=90, font=\small\bfseries] at (-1.3, 2.4) {Queries $\mathbf{Q}$};

\fill[vir100]  (5.6, 3.84) rectangle (6.1, 4.80);
\fill[vir065]  (5.6, 2.88) rectangle (6.1, 3.84);
\fill[vir035]  (5.6, 1.92) rectangle (6.1, 2.88);
\fill[vir010]  (5.6, 0.96) rectangle (6.1, 1.92);
\fill[vir005]  (5.6, 0.00) rectangle (6.1, 0.96);
\draw[black!50, thin] (5.6, 0.0) rectangle (6.1, 4.8);
\node[right, font=\footnotesize] at (6.1, 4.56) {1.0};
\node[right, font=\footnotesize] at (6.1, 2.40) {0.5};
\node[right, font=\footnotesize] at (6.1, 0.24) {0.0};
\node[above, font=\footnotesize, align=center] at (5.85, 4.8) {Attn.\\wt.};
\fill[virMask] (5.6, -0.8) rectangle (6.1, -0.2);
\draw[black!50, thin] (5.6, -0.8) rectangle (6.1, -0.2);
\node[right, font=\scriptsize] at (6.1, -0.5) {masked};

\draw[orange!70!black, very thick, rounded corners=1pt]
  (0, 0) rectangle (4.8, 1.2);
\node[black!70!black, font=\footnotesize, align=left, below=0.15cm]
  at (1, 0)
  {``sweet'' attends most strongly to ``apple'' (0.65)
   $\Rightarrow$ low-weight\\ KV pairs $\to$ eviction candidates};

\end{tikzpicture}
\caption{Causal self-attention weight matrix for ``The apple tastes sweet.'' visualised
  with the Viridis colormap (dark purple = low, yellow = high). Gray cells are
  causally masked future tokens. Each row sums to 1 (post-softmax). Query
  ``sweet'' concentrates 65\% of its attention on ``apple'', demonstrating that KV entries carry highly non-uniform importance, the core premise of
  attention-score-driven eviction methods such as H$_2$O and SnapKV.}
\label{fig:attn_heatmap}
\end{figure}

This process repeats for each new token during generation, enabling the model to incorporate contextual information efficiently. Fig.~\ref{fig:attn_heatmap} provides a concrete numerical example of Eqs. (3)--(5) for the sentence ``The apple tastes sweet,'' highlighting the non-uniform attention distribution that motivates
selective KV cache eviction.

\

\begin{figure}[htbp]
\centering
\begin{tikzpicture}[
  root/.style={rectangle, draw, fill=teal!25, rounded corners=4pt,
               minimum width=4.5cm, minimum height=0.9cm,
               font=\small\bfseries, align=center},
  cat/.style={rectangle, draw, fill=orange!20, rounded corners=3pt,
              minimum width=3cm, minimum height=0.8cm,
              font=\small, align=center},
  arrow/.style={-{Stealth}, thick},
  level 1/.style={sibling distance=3.4cm, level distance=2.2cm}
]
\node[root] (root) {KV Cache\\Optimization};

\node[cat] (evict)  at (-6.8,-2.2) {Cache\\Eviction};
\node[cat] (comp)   at (-3.4,-2.2) {Cache\\Compression};
\node[cat] (hybrid) at (0,-2.2)    {Hybrid\\Memory};
\node[cat] (attn)   at (3.4,-2.2)  {New Attention\\Mechanism};
\node[cat] (comb)   at (6.8,-2.2)  {Combination\\Methods};

\foreach \n in {evict, comp, hybrid, attn, comb}
    \draw[arrow] (root) -- (\n);

\node[font=\tiny, align=center, below=0.15cm of evict]
    {H$_2$O, SnapKV,\\NACL, Ada-KV\\\ldots};
\node[font=\tiny, align=center, below=0.15cm of comp]
    {KIVI, PALU,\\MiniCache,\\KVQuant};
\node[font=\tiny, align=center, below=0.15cm of hybrid]
    {PagedAttention,\\InfiniGen,\\LayerKV \ldots};
\node[font=\tiny, align=center, below=0.15cm of attn]
    {Linear,\\Log-Linear,\\KIMI Linear};
\node[font=\tiny, align=center, below=0.15cm of comb]
    {FlexGen,\\ShadowKV,\\TailorKV};

\end{tikzpicture}
\caption{Taxonomy of KV cache optimization techniques surveyed in this paper, organized into five major categories.}
\label{fig:taxonomy}
\end{figure}
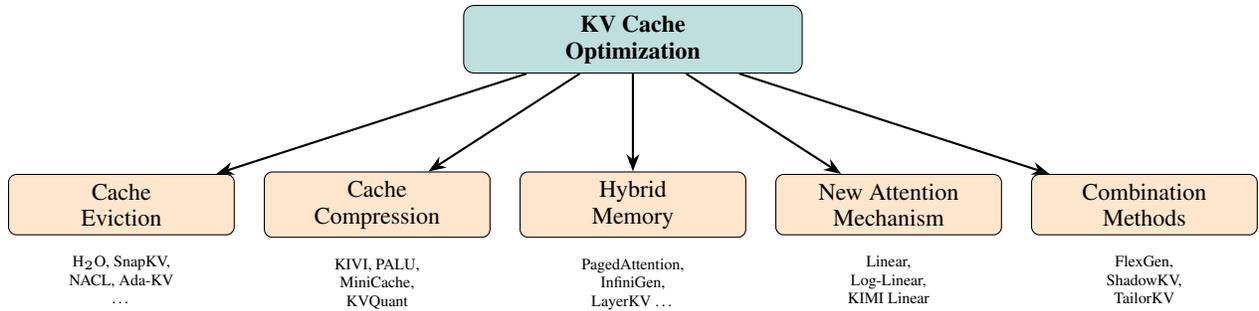

\subsection{Taxonomy}
To provide a structured overview, we present a taxonomy that groups existing methods into five major categories shown in Fig.~\ref{fig:taxonomy}. Each category is characterized by the system-level objectives it aims to improve alongside the trade-offs it introduces Table~\ref{tab:kv_cache_optimization}. This taxonomy offers readers a high-level comparison of design motivations and limitations across approaches, serving as a reference point for understanding how different methods prioritize performance, efficiency, and model quality.


\begin{table}[!t]
	\caption{Comparison of KV Cache optimization techniques}
	\centering
    \begin{tabular*}{\textwidth}{@{\extracolsep{\fill}}
  >{\raggedright\arraybackslash}p{0.15\textwidth}
  >{\raggedright\arraybackslash}p{0.18\textwidth}
  >{\raggedright\arraybackslash}p{0.18\textwidth}
  >{\raggedright\arraybackslash}p{0.18\textwidth}
  >{\raggedright\arraybackslash}p{0.20\textwidth}@{}
}
    \toprule
		\textbf{Technique} & \textbf{Optimization Goal} & \textbf{Tradeoffs} & \textbf{Representative Methods} & \textbf{Good for} \\
		\midrule
		Cache Eviction & Memory Footprint, Decoding Latency & Accuracy and quality & H$_2$O, SnapKV, NACL, InfiniPot, Hashevict, MorphKV, RocketKV, KVzip, Ada-KV & Long-context single requests, Edge devices, Minimal model modification  \\
		Cache Compression & KV cache size, Throughput & Dequantization and reconstruction overhead, Accuracy & KIVI, MiniCache, PALU, KVQuant & Bandwidth-bound workloads, Edge devices, Ultra-long contexts  \\
		Hybrid memory solution & TTFT in long-context, System efficiency & Hardware requirement, complexity in management & PagedAttention, InfiniGen, LayerKV, INF2, KVPR, Oneiros, CLO & High‑throughput serving, Multi-tenant datacenter serving  \\
		New Attention Calculation & computational complexity, Inference Speed & Accuracy, model retraining  & Linear, Log-Linear, Local Linear, KIMI Linear & Agentic/Long-horizon Tasks \\
		Combination Methods & Throughput/Latency Balance & Complexity in design and management & FlexGen, Q-Hitter, ShadowKV, TailorKV & Consumer-Grade Hardware and Diverse Workloads, balanced performance \\
		\bottomrule
	\end{tabular*}
	\label{tab:kv_cache_optimization}
\end{table}

\section{Content}
As discussed earlier, increasing context length puts significant pressure on GPU memory and memory bandwidth, thus slowing down token throughput. While modern LLMs continue to expand their context window sizes to support more complex tasks, recent models have reached millions of tokens. \cite{GPT_Model_Compare, Llama4} This trend further amplifies the memory and computational challenges associated with KV cache management. 

To address these challenges, both academia and industry have proposed a range of optimization strategies. This survey reviews and categorizes these approaches into five major directions: (1) cache eviction methods that selectively discard less critical tokens, (2) compression and reconstruction techniques that reduce memory footprint, (3) hybrid memory solutions leveraging multi-tier storage, (4) novel attention mechanisms that rethink context processing, and (5) combination strategies that integrate multiple optimizations.


\subsection{Cache Eviction}
Cache eviction is a strategy that selectively discards the less important cache while retaining the critical ones for accurate generation. This method usually relies on attention scores to estimate the token importance. Tokens with high scores are retained while those with low scores are discarded. This method maintains KV cache in a manageable size to reduce memory footprint. The summary of all the methods described in this section is given in Table~\ref{tab:kv_cache_eviction}. 

\begin{figure}[!tb]
    \centering
    \includegraphics[width=0.9\linewidth]{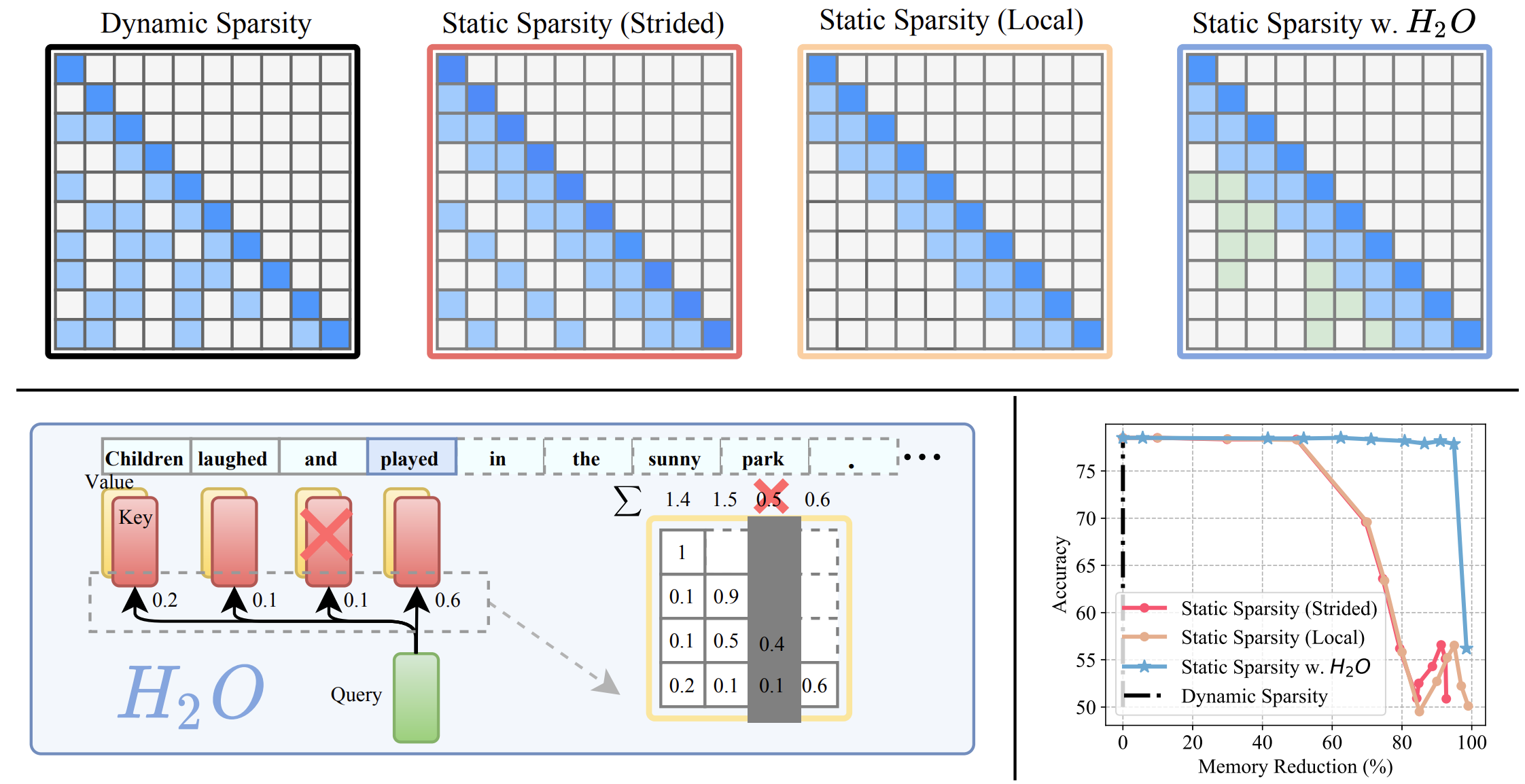}
    \caption{Upper plots illustrate symbolic plots of an attention map deploying different KV cache policies in
LLM generation. Lower right: contrasts their accuracy-memory trade-off. Left: the overview of H$_2$O framework~\cite{H2O}.}
    \label{fig:h2o}
\end{figure}

A key challenge in eviction-based methods is identifying which tokens carry long‑range importance. One approach tracks accumulated attention scores and treats high‑scoring tokens as essential for maintaining model accuracy. Work by Zhengyu et al.~\cite{H2O} shown in Fig.~\ref{fig:h2o} denotes the tokens with high accumulated attention scores as heavy-hitter (H2) tokens. Losing the H2 tokens leads to significant performance degradation during generation. To address this, the authors propose H$_2$O (Heavy-Hitter Oracle), a dynamic eviction policy that balances retention of recent tokens with heavy-hitter tokens. The researchers used a greedy algorithm to manage a fixed cache size budget. The algorithm starts with an empty cache. For each new token, if the cache has space, the token will be added to the cache. Otherwise, the system computes the accumulated attention scores for each token in the cache, including the new token. The token with the lowest score will be discarded to maintain the cache size. 

Another method addresses token selection during the encoding phase. SnapKV \cite{SnapKV} introduced a token selection and compression strategy designed for long-context prompts. SnapKV applies a voting mechanism within the observation window to identify the most important previous tokens. An observation window is the most recent segment of tokens in the input. It is used to calculate attention scores for each previous token. The scores are subsequently aggregated to identify the most influential tokens across the entire sequence. To preserve contextual coherence, SnapKV employs 1D pooling to cluster nearby important tokens to ``retain the features surrounding the selected attention features''~\cite{SnapKV}. In other words, SnapKV retains not only the top K important tokens, but also the context surrounding them. During the compression phase, SnapKV keeps both the selected important tokens from the prefix and all tokens in the observation window. This approach enables efficient KV cache compression for long prompts while maintaining high accuracy. 

\begin{figure}[!tb]
    \centering
    \includegraphics[width=0.9\linewidth]{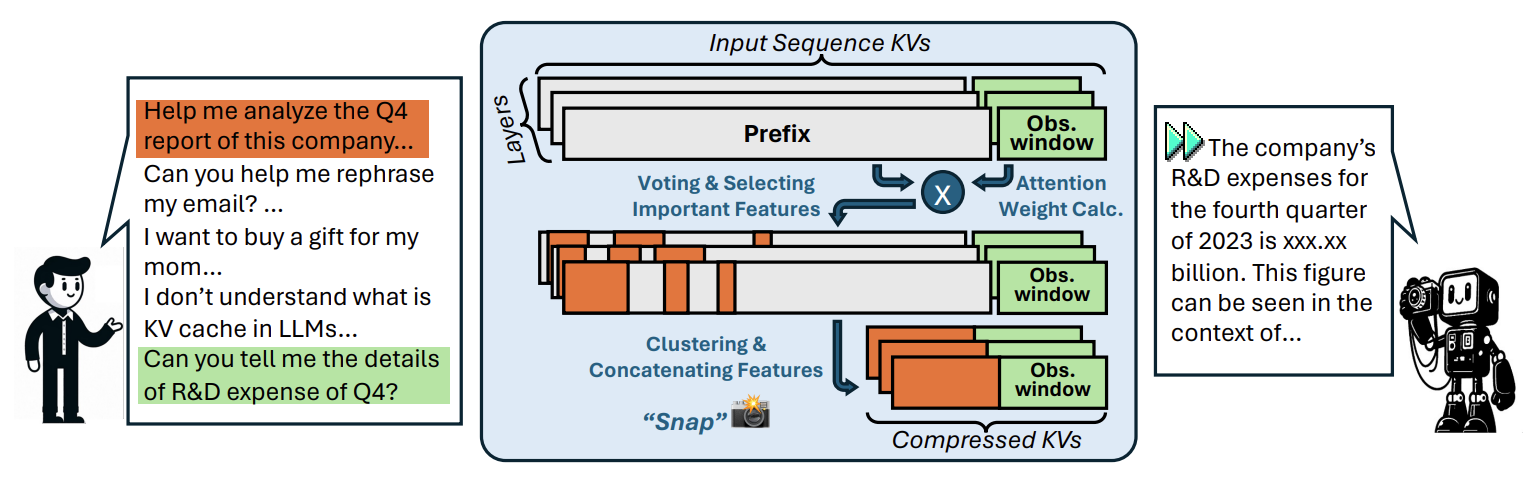}
    \caption{The graph shows the simplified workflow of SnapKV, where the orange area represents the cluster of features per head selected by SnapKV. These features are then used to form new Key-Value pairs concatenated with the features in the observation window. Together, the selected prefix and observation windows constitute the new KV cache utilized for the generation.~\cite{SnapKV}.}
    \label{fig:Snap_KV}
\end{figure}

In contrast to step‑wise eviction, the work in NACL \cite{NACL} adopts a single-shot KV cache eviction strategy that combines Proxy-Tokens Eviction with Random Eviction. Proxy tokens are a small subset within the input, that ``are responsible for yielding the most precise outcomes during the computation of the token score'' \cite{NACL}. Usually, NACL picks the few tokens at the end of the input, such as the user’s question, as the proxy tokens. A scoring function calculates the accumulated attention scores of all other tokens against only these selected proxy tokens, providing a precise measurement of tokens’ importance in the task-relevant context. The less imporant tokens will be discarded. 
To further optimize cache size, NACL incorporates Random Eviction. The token selection is based on a probability distribution, rather than pure randomness. The attention scores calculated between proxy tokens and all other tokens are turned into a probability distribution using SoftMax. Tokens are then sampled from this distribution to fill the remaining eviction budget. The selected tokens are discarded all-together during encoding, leading to one big eviction, rather than one by on. Compared to traditional methods (like H$_2$O introduced in previous section) that discard tokens one by one repeatedly, this approach reduces computation stress. 

Some recent research explores continual distillation as a strategy for managing long contexts. The approach in \cite{InfiniPot} uses novelty measurements to retain essential information when memory limits are reached. The authors take an analogy of InifiniPot to be like a cooking pot: when a pot nears overflow, it distills unnecessary parts and retains only the essentials. InfiniPot enables pre-trained LLMs to handle ``infinitely'' long contexts within fixed memory constraints by introducing Continual Context Distillation (CCD). Token importance is determined with two metrics: CaP (Catalyst Prompt) and NuC (Novelty under Compression). CaP measures how representative a token is for future context. The system adds a short prompt like “Summarize the critical points” before the pot overflows and then computes attention scores to see which tokens matter most. NuC measures how new or unique a token is compared to existing ones. Tokens that are harder to predict will get higher NuC scores. Before memory saturation, the system triggers a distillation step. Based on the CaP and NuC scores, the system keeps the top K tokens and distills the rest. This dynamic distillation ensures that critical and novel information remains accessible while maintaining a constant memory footprint. 

A lightweight alternative avoids full attention computation altogether. HASHEVICT \cite{HashEvict} introduces a pre-attention KV cache eviction strategy that leverages Locality-Sensitive Hashing (LSH) to estimate token similarity before performing full attention computation, thereby reducing both memory and computational overhead. Specifically, it uses SimHash, an LSH algorithm where similar data points produce similar hash codes. In HASHEVICT, each token’s key embedding and the current token’s query embedding is converted into a binary hash code using random projections. Token similarity is then approximated by computing the Hamming distance between these hash codes. As shown in Section 2.9.2 of~\cite{hamming}, hamming distance is the number of positions where two codewords of the same length differ. Here it refers to the number of differing bits between the two binary hash codes. Tokens with the largest Hamming distance (least similarity) are assumed to have low attention relevance and are evicted.  
This method is efficient and lightweight because the operations used for hashing and hamming distance calculation are simple and GPU friendly. Therefore, by enabling a single round of token eviction prior to the actual attention calculation, HASHEVICT helps to reduce both memory and computational cost. 

\begin{table}[!t]
	\caption{Summary of KV Cache eviction techniques}
	\centering
\begin{tabular*}{\textwidth}{@{\extracolsep{\fill}}p{2.5cm}p{3cm}p{1.7cm}p{8cm}@{}}
\toprule
\textbf{Method} & \textbf{Mechanism} & \textbf{Phase} & \textbf{Overview} \\
\midrule
H$_2$O~\cite{H2O} &
Evict top K noncritical tokens &
Decoding &
Retains a balance of Heavy‑Hitter tokens (H2) and the most recent tokens \\
\addlinespace
SnapKV~\cite{SnapKV} &
Retain critical tokens and their context &
After Prefill &
Identifies critical attention features by voting from observation window and uses pooling for clustering \\
\addlinespace
NACL~\cite{NACL} &
Hybrid Proxy and Random Eviction &
Prefill &
Performs a single, global eviction during encoding \\
\addlinespace
InfiniPot~\cite{InfiniPot} &
Continual Context Distillation &
Prefill &
Enables processing of infinite context within a fixed memory budget \\
\addlinespace
HASHEVICT~\cite{HashEvict} &
Attention‑Free LSH Ranking &
Decoding &
Uses Locality‑Sensitive Hashing (LSH) and Hamming Distance to estimate token importance \\
\addlinespace
MorphKV~\cite{MorphKV} &
Correlation‑aware selection &
Decoding &
Maintains a constant‑sized cache for long‑response tasks while eliminating early‑token bias \\
\addlinespace
RocketKV~\cite{RocketKV} &
Two‑stage compression &
Prefill and Decoding &
Two‑stage compression: permanent coarse eviction (as in SnapKV) followed by fine‑grain dynamic selection (HSA) \\
\addlinespace
KVzip~\cite{KVzip} &
Context reconstruction &
Prefill &
Importance scoring using maximum cross‑attention \\
\addlinespace
Ada‑KV~\cite{Ada-KV} &
Adaptive Budget Allocation &
After Prefill &
Dynamically allocates cache budget across attention heads rather than a fixed budget per head \\
\bottomrule
\end{tabular*}
	\label{tab:kv_cache_eviction}
\end{table}

Another technique focuses on leveraging recent attention patterns. The framework presented in MorphKV~\cite{MorphKV} dynamically selects relevant old tokens based on aggregated or peak attention signals. To be more specific, it uses attention patterns of recent tokens to decide which older tokens remain relevant. MorphKV only preserves the most relevant old tokens and a fixed number of recent tokens to limit KV cache size. The system selects the important old tokens by scoring for each old token with either Sum Fusion or Max Fusion function. Both of them are based on Softmax attention scores.   
Sum Fusion aggregates attention scores across recent tokens, favoring tokens consistently attended by multiple recent tokens. It is ideal for maintaining contextual consistency. In contrast, Max Fusion selects tokens based on the highest individual attention score, making it suitable for tasks requiring sharp, focused retrieval. If at least one recent token strongly cares about a particular old token, it implies that the old token is of high importance under Max fusion. Therefore, by dynamically selecting tokens based on these relevance measures, MorphKV is able to keep a constant KV cache size while preserving accuracy. 

A two-stage design is explored in RocketKV \cite{RocketKV}, which combines coarse eviction with dynamic sparse selection to accelerate long-context inference without significant accuracy loss. RocketKV uses SnapKV \cite{RocketKV, SnapKV} to do coarse-grain KV cache eviction in the first stage and uses Hybrid Sparse Attention (HAS) to select relevant tokens dynamically in the second stage. HSA groups tokens into pages and stores the maximum and minimum key values for each page, enabling fast approximation of token importance page wise. Based on these approximations, the system retrieves the original KV pairs for the selected pages and uses them for attention computation. This hierarchical approach balances efficiency and accuracy, significantly improving throughput for large-context LLMs. 

A reconstruction-driven and query-agnostic KV cache eviction approach is introduced in KVzip \cite{KVzip}, where token importance is derived from a self‑supervised context reconstruction process rather than from query‑specific signals. During the initial context encoding, the model processes a full input context and generates a complete KV cache. To assess token significance, KVzip performs a self-supervised reconstruction task by prompting the model with instructions such as “Repeat the previous content” followed by the original context. This reconstruction simulates an autoencoder-like mechanism, where the model attempts to regenerate the input using only the cached KV pairs. During this process, the model's attention mechanism naturally reveals the importance of KV pairs. Tokens that contribute most to accurate reconstruction are deemed critical, while those with lower impact are evicted. This approach effectively identifies essential KV pairs without relying on query-specific information, enabling efficient compression while preserving contextual integrity. 

In traditional cache eviction methods, there exists a key limitation; that is the fixed cache budget is distributed uniformly across all attention heads while ignoring head-specific attention patterns. To address this issue, Yuan et al. proposed Ada-KV \cite{Ada-KV}, a head-wise adaptive budget allocation strategy. The core idea of this Ada-KV approach is to minimize the eviction loss by reallocating budget from ``attention-sparse heads'' (where weights are concentrated on few elements) to ``attention-dispersed heads'' (with widespread concentration patterns). The authors establish a theoretical framework for cache eviction by defining eviction loss and deriving its upper bound \cite{Ada-KV}. Each attention head dynamically retains a variable number of critical tokens based on its attention distribution, rather than a fixed amount to every head like the traditional methods do.  Furthermore, Ada-KV is designed as a plug-and-play solution that can seamlessly integrate with other methods, such as SnapKV, to enhance overall efficiency. 

\subsection{Cache Compression}
Cache compression aims to reduce memory footprint. The most common approach is quantization, which represents numerical values with fewer bits. The common format for modern transformer models to store activations during inference is fp16, also known as 16-bit floating point \cite{16-bit}. Quantization can reduce the size to a 4-bit integer \cite{4-bit, Qrazor}. Along with cache quantization, other compression strategies target structural redundancies, such as layer-wise or matrix-level compression, enabling further optimization of storage and computational efficiency as given in the summary form in Table~\ref{tab:kv_comp}.

One line of compression research focuses on asymmetric quantization. The method proposed in KIVI \cite{KIVI} applies per‑channel and per‑token quantization schemes to preserve precision while reducing memory. KIVI stands for Key-Value cache with Input-dependent quantization. The authors observe that for key cache, “there are a few fixed channels whose magnitudes are very large”; whereas for value cache, “there is no obvious outlier pattern”. Based on this insight, KIVI applies per-channel quantization for keys and per-token quantization for values, preserving precision where it matters most. To illustrate, if one channel contains large values and another contains small values, quantizing them separately maintains accuracy compared to uniform quantization. For example, if we have 3 tokens and 2 channels as: 

\begin{center}
\texttt{Token1: [10, 0.5] \qquad Token2: [12, 0.6] \qquad Token3: [15, 0.5]}   
\end{center}
 
Since the first channel has large values and the second channel has small values, we can quantize the values based on different channels to keep each channel’s precision. That gives us the first channel (10, 12, 15)  to be 0, 1, 3, and the second channel (0.5, 0.6, 0.5) to be 0, 3, 0.  

\begin{figure}[!tb]
    \centering
    \includegraphics[width=0.4\linewidth]{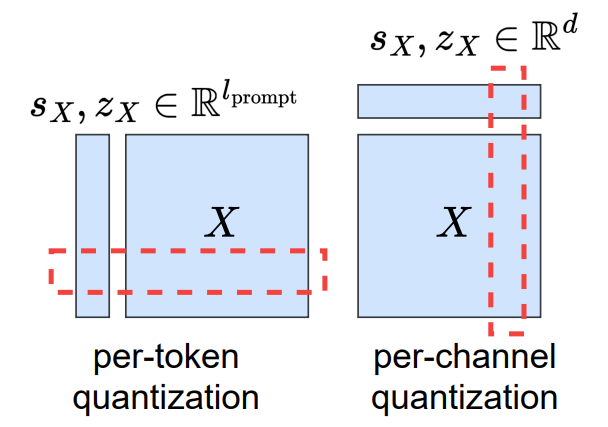}
    \caption{Definition of per-token and per-channel quantization. 
$X \in \mathbb{R}^{l_{\text{prompt}} \times d}$ is the key/value cache, where $l_{\text{prompt}}$ is the number of tokens and $d$ is the number of channels. 
$z_X$ is the zero-point, and $s_X$ is the scaling factor..~\cite{KIVI}.}
    \label{fig:KIVI}
\end{figure}

Additionally, KIVI organizes the KV cache into groups (e.g., 32 tokens per group) and residuals (recent tokens kept in full precision).  The group will be quantized to save memory, while the residuals will be kept in full precision for accuracy. As generation progresses, residuals are merged into groups after quantization, ensuring a balance between efficiency and performance. For key cache, this happens every time the number of residuals reaches the pre-defined threshold length. For value cache, this happens every time a new token is generated. This dynamic grouping mechanism allows KIVI to achieve significant memory savings without compromising inference quality. 

\begin{figure}[!b]
    \centering
    \includegraphics[width=0.5\linewidth]{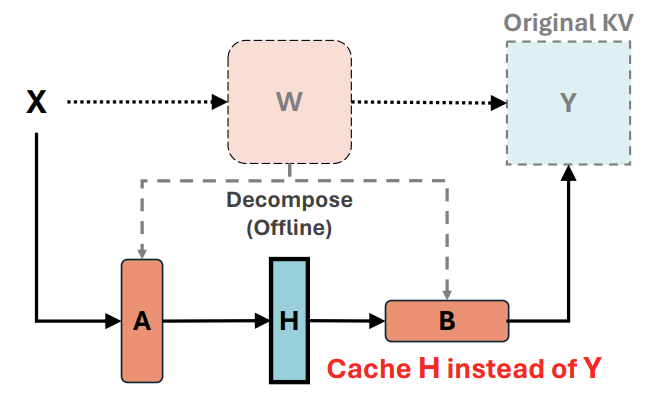}
    \caption{Palu's low-rank projection method for KV-cache reduction. 
    A weight matrix $\mathbf{W}$ of linear projection is decomposed into two low-rank matrices. 
    Input $\mathbf{X}$ is down-projected to a latent representation $\mathbf{H}$, which is cached. 
    $\mathbf{Y}$ can be reconstructed from $\mathbf{H}$ using the up-projection matrix $\mathbf{B}$.~\cite{PALU}.}
    \label{fig:PALU}
\end{figure}

While most of the conventional methods focus on redundancy within a single layer, MiniCache introduces a cross-layer KV cache compression technique that exploits redundancy between adjacent layers. The researchers observed that KV cache states show high similarity between adjacent layers in the middle to deep layers, as the representation start to stabilize in those layers.  
With this observation, this method begins by measuring similarity between KV states of neighboring layers using angular distance (or cosine similarity). If the similarity falls below a predefined threshold, the two layers’ KV states are considered mergeable. To preserve information during merging, MiniCache employs SLERP (Spherical Linear Interpolation) rather than direct averaging. After merging, it stores the interpolated direction, the original magnitudes of both original vectors, and the angle between the vectors, enabling accurate reconstruction when needed by rescaling the stored direction. This approach significantly reduces memory usage while maintaining the ability to restore original KV states for attention computation. 

Low‑rank projection has also been explored to compress the KV cache. With the system described in PALU \cite{PALU}, each projection matrix W for both keys and values are decomposed into two smaller matrices A and B Via Singular Value Decomposition (SVD), such that W $\approx$ A $\times$ B. For example, if the input token vector X has a size of 4096 and we know each attention head uses 128 dimensions, the projection matrix W will have a size of 4096 $\times$ 128. Matrix W can be factorized into matrix A (4096$\times$r) and matrix B (r$\times$128), where r is a smaller value than 128, for example, 32. SVD mathematically finds the best low-rank approximation of W, meaning that A x B is close to the original W.

During inference, the system multiplies input X (size = 4096) by A (size = 4096 x r) to obtain a latent representation H (size r). Instead of caching the original KV, only the small sized H is cached. When reconstruction is needed, multiplying H by B restores the original KV representation. 
To minimize runtime overhead, Palu uses Matrix Fusion to optimize computing costs during reconstruction. Palu fuses the reconstruction matrix (matrix B) into other existing weight matrices offline before inference, avoiding separate reconstruction steps. However, for Keys that use Rotational Positional Embedding (RoPE), matrix fusion is not possible, so PALU reconstructs the Key dynamically using a custom and highly efficient GPU kernel. 
Group-Head Low-Rank Decomposition (G-LRD) is another optimization method used. Joint decomposition across all heads provides high accuracy, yet per-head decomposition allows low reconstruction cost. To balance accuracy and cost, PALU groups several attention heads together and performs decomposition on that group.

Ultra‑low‑bit quantization is taken further in KVQuant \cite{KVQuant}, which integrates per‑channel quantization, pre‑RoPE processing, non‑uniform quantization, and outlier preservation to enable LLM inference with context lengths up to 10 million tokens. The method combines the following key components:  
\begin{table}[!pb]
\centering
\caption{Cache Compression Methods Comparison Table}
\label{tab:kv_comp}
\begin{tabularx}{\textwidth}{@{}p{2cm}p{4cm}p{3.5cm}X@{}}
\toprule
\textbf{Method} & \textbf{Mechanism} & \textbf{Granularity} & \textbf{Outlier handle} \\
\midrule
KIVI~\cite{KIVI} &
Asymmetric Quantization &
\textbf{Key}: Per-channel; \textbf{Value}: Per-token &
Keeps a small amount of residual cache in full precision \\
\addlinespace
KVQuant~\cite{KVQuant} &
Non-Uniform Quantization (NUQ) &
\textbf{Key}: Per-channel (Pre-RoPE); \textbf{Value}: Per-token &
Isolates the top 1\% of outliers into a separate sparse fp16 representation \\
\addlinespace
MiniCache~\cite{MiniCache} &
Cross-Layer KV Merge &
NA &
Keeps highly distinct unmergeable state pairs in full precision \\
\addlinespace
PALU~\cite{PALU} &
Hidden-dimension compression with Low-Rank Projection and Reconstruction &
Per-token per head group latent vector &
Assigns higher ranks to critical layers \\
\bottomrule
\end{tabularx}
\end{table}
\paragraph{Per-Channel Key Quantization:} Similar to KIVI, keys are quantized per channel, instead of per token, to save memory and improve accuracy.

\paragraph{Pre-RoPE Quantization:} RopE is a rotation that LLMs apply to keys. This rotation makes quantization harder. Pre-RoPE Key Quantization applies quantization on keys before RoPE rotation is applied. This method simplifies computation and maintains structural integrity for higher accuracy.  

\paragraph{Sensitivity-Weighted Non-Uniform Quantization:} Compared to traditional uniform quantization that evenly spaces the quantization levels, KVQuant uses Sensitivity-Weighted Non-Uniform Quantization, where the quantization levels are optimized based on the data distribution to handle the outliers effectively. For example, for a key vector [0.02, 0.03, 0.05, 0.07, 10.0]. For uniform quantization into 3 bits (2$^{3}$ --> 8 levels), we evenly space the quantization levels to get levels: [-10, -7.5, -5, -2.5, 0, 2.5, 5, 7.5, 10]. In this case, all small numbers (0.02–0.07) map to 0, losing precision.  In contrast, using non-uniform quantization gives us [0, 0.02, 0.04, 0.06, 0.08, 1, 5, 10]. Now small numbers keep details, and big outliers still fit.

\paragraph{Per-Vector Dense-and-Sparse Quantization}: Extreme outliers (~1\% of elements) are stored separately in higher precision, while the majority of values are quantized to low bit, balancing memory savings and accuracy. 

\paragraph{Attention Sink-Aware Quantization}: The first token is stored in full FP16 precision because LLM tends to allocate a large attention score to the first token, even when the initial token is semantically less important. 

By combining these techniques, KVQuant achieves aggressive compression while preserving model fidelity, making ultra-long context inference feasible.

\subsection{Hybrid Memory Solution}

Hybrid memory solutions address the limitations of GPU memory by leveraging multi-tier storage architectures to manage the KV cache efficiently. As context lengths grow, storing the entire KV cache on GPU becomes infeasible due to memory constraints and bandwidth bottlenecks. Hybrid approaches mitigate these challenges by offloading portions of the cache to slower but larger memory tiers, such as CPU memory, disk, or specialized accelerators. Hybrid memory solution aims to balance memory availability and computational efficiency. The solutions discussed in this section are summarized in Table~\ref{tab:hybrid_memory}.

A widely adopted hybrid memory strategy begins with treating KV storage like paged virtual memory. PagedAttention \cite{PagedAttention} is an attention algorithm ``inspired by the operating system's (OS) solution to memory fragmentation and sharing: virtual memory with paging'' \cite{PagedAttention}. It divides the KV cache into fixed-size units called KV blocks, analogous to memory pages. These blocks do not need to be stored next to each other in the physical GPU memory. A block table maps logical blocks into physical locations, similar to page tables in OS. During attention computation, the model retrieves only the required blocks, operating block-by-block rather than assuming contiguous storage. Such block-level management enables efficient memory sharing. 
This design facilitates efficient memory sharing, particularly in complex decoding scenarios such as parallel sampling, where multiple sequences share the same input prompt, and the KV blocks corresponding to the prompt are physically shared. To maintain consistency when a shared block is modified, PagedAttention employs a copy-on-write mechanism by duplicating only the affected block instead of the entire cache. This approach significantly reduces memory overhead while supporting scalable multi-sequence inference. 

\begin{figure}[!b]
    \centering
    \includegraphics[width=0.6\linewidth]{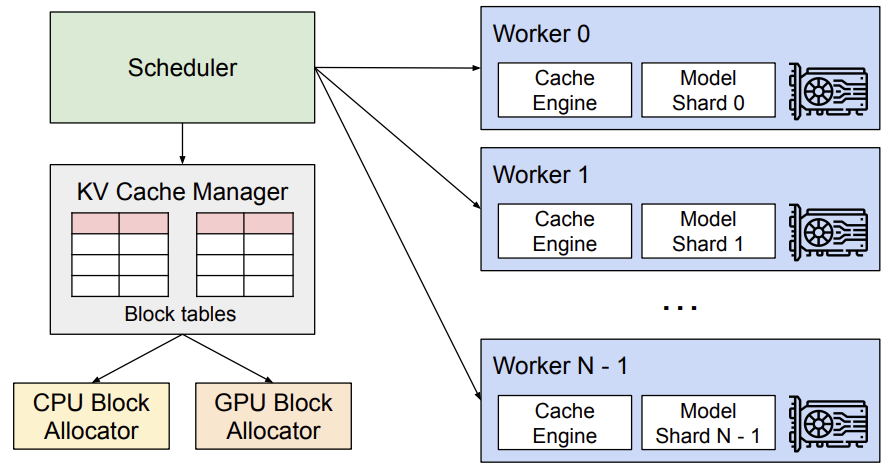}
    \caption{vLLM system overview~\cite{PagedAttention}.}
    \label{fig:PagedAttention}
\end{figure}

Another hybrid memory solution predicts future KV needs during inference. InfiniGen \cite{InfiniGen} introduces a dynamic KV cache management strategy that stores the cache in CPU memory and selectively transfers only the most critical KV pairs to the GPU for attention computation, minimizing data transfer overhead. Its key innovation lies in predictive prefetching, that is anticipating which tokens will be needed before they are actually required, enabling efficient prefetching. 

To achieve this, InfiniGen leverages the high similarity between adjacent query vectors to calculate an approximate query vector, known as Partial Q. This approximation process is a single matrix multiplication using current layer input and the next layer’s query weight $W_Q^{(\text{layer}+1)}$, which is pre-loaded. The accuracy of $\tilde Q$ is further improved by applying a low-rank transformation M learned offline via SVD. 

\begin{equation}
\tilde Q^{(\text{layer}+1)} = X^{(\text{layer})} \cdot M \cdot W_Q^{(\text{layer}+1)}
\end{equation}

\begin{figure}[!tb]
    \centering
    \includegraphics[width=0.6\linewidth]{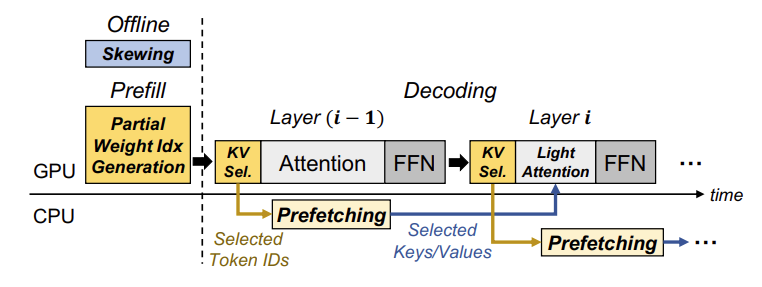}
    \caption{Operation flow of the prefetching module of InfiniGen. ~\cite{InfiniGen}.}
    \label{fig:InfiniGen}
\end{figure}

Using $\tilde Q$ and KV metadata, InifiniGen predicts which KVs could be critical to the next layer, and prefetch them from CPU to GPU while the GPU is still processing the current layer. This overlapping of computation and data transfer significantly improves throughput for long-context inference. 

A layer‑wise KV cache management strategy is proposed in LayerKV \cite{LayerKV}. The core concept is to split KV cache by layers, keeping only a subset of layers on the GPU during the prefill stage while offloading some layers to CPU memory to reduce Time to First Token (TTFT). 
Prefill time refers to the time for the GPU to compute the first token. Offload time is the time required for the CPU to transfer KV data back to the GPU. To hide transfer latency, LayerKV selects the minimum number of layers to keep on the GPU such that: offload time <= prefill time. For example, in an 8‑layer model, LayerKV may keep layers 1, 3, 5, and 7 on the GPU while offloading layers 0, 2, 4, and 6 to CPU memory. As the GPU processes layer 1, the CPU simultaneously transfers KV data for layer 0 back to the GPU, creating overlap between computation and communication. This overlap significantly reduces TTFT. 
Additionally, LayerKV integrates an SLO-aware scheduler that determines how many new inference requests can be initiated without violating Time Per Output Token (TPOT) guarantees. This ensures that both latency and throughput targets are met under multi-request serving scenarios. 

A hardware–software co‑design is explored in INF2 \cite{INF2}. INF2 stands for ``INFerence-INFinity''. This method leverages Computational Storage Devices (CSDs), which are special SSDs with attached accelerators like FPGA. The key mechanism is to offload attention computation to accelerators near storage.  
During the decoding phase, instead of transferring large KV caches back and forth between GPU and storage, INF2 stores the KV cache directly on the SSDs within the CSDs. The KV data is loaded into the attached accelerator via a private PCIe switch, allowing multi-head attention to be computed locally on the accelerator, close to where the KV resides. While the accelerator performs the attention computation near storage, the GPU concurrently processes the remaining computation like MLP. Newly generated KV entries are temporarily stored in RAM and will be sent to SSD in batches. By relocating attention computation to near-storage accelerators and reducing PCIe traffic, INF2 significantly enhances LLM throughput for long-context inference and alleviates pressure on GPU memory bandwidth. 

Another system aims to reduce GPU idle time. The technique in KVPR \cite{KVPR} overlaps partial KV reconputation on GPU with data transfer from CPU to GPU, synchronizing both to minimize stalls. Instead of waiting for the full KV cache to arrive, KVPR first transfers a small portion of the input activations, enabling the GPU to immediately begin recomputing a corresponding subset of the KV cache. Meanwhile, the system continues to transfer the rest of the KV cache in parallel.  

In a profiling stage, the system measures key parameters such as PCIe bandwidth, GPU compute throughput, and model sizes. Subsequently, the system determines whether the deployment target prioritizes low latency or high throughput. Based on this objective, a scheduling module selects between two processing strategies: row-wise scheduling for latency-sensitive workloads, or column-wise scheduling for throughput-oriented workloads. Using the profiling results, KVPR computes the optimal amount of KV cache to recompute. This calculation is performed using the parameters collected from the prefiling phase, where the small amount of KV cache’s recomputation time matches the transfer time for the remaining KV cache. This creates a tightly synchronized pipeline where GPU computation and data transfer overlap almost perfectly. During inference, the system first sends the activations needed for recomputation. The GPU begins partial KV regeneration as soon as they arrive, while the CPU continues streaming the reamining KV cache. Once both the recomputing and transferring phases finish, the recomputed and transferred KV cache segments are merged for the subsequent attention calculation. 

A multi‑tenant serving strategy is demonstrated in Oneiros \cite{Oneiros}, which temporarily remaps model parameters off GPU to free memory for KV cache expansion during heavy decoding workloads. It is designed for multi-tenant LLM serving on modern hardware with high bandwidth. It reuses GPU memory allocated for model parameters as extra space for KV cache, a process called “parameter remapping”. To be more specific, when the KV cache approaches the GPU’s memory limit, Oneiros triggers remapping to free space.  

The system first identifies which parameters to offload for remapping. Normally, inactive models are the preferred candidates as their parameters are unlikely to be needed immediately. If all models are active, then the system will offload a subset of layers from each model in an evenly spaced manner.  
This strategy takes advantage of the sequential layer-by-layer execution inherent in autoregressive decoding. Since LLM inference repeatedly cycles through layers for each token, distributing remapped layers evenly ensures sufficient compute time between remapped layers, allowing parameter transfers from CPU to GPU to be effectively hidden behind ongoing computation.  
After offloading the inactive models or layers, the newly released GPU memory becomes available to store additional KV cache entries. If a remapped layer is later required, its parameters are fetched back from CPU to GPU memory, while the GPU continues computing the next available layers, overlapping communication with computation to minimize latency impact. When KV cache pressure subsides, Oneiros restores GPU memory to model parameters, reversing the remapping process as needed. 

To reduce redundant KV transfers, Jiawei et al. proposed CLO~\cite{CLO}. It is a KV‑cache offloading and reuse mechanism designed to accelerate LLM inference by intelligently moving KV cache between CPU and GPU memory. Its key insight is that Query vectors generated during adjacent decoding steps (generating one token after another) are often highly similar. This high temporal locality suggests that the corresponding KV entries needed for attention are also likely to be the same. CLO measures the cosine similarity between the current query and the previous one. If the similarity exceeds a threshold, the system assumes a high likelihood of accessing the same KV vectors and therefore reuses the previously loaded KV cache instead of fetching new KV data from CPU memory. This significantly reduces cache movement overhead.  When similarity is low, CLO must fetch the correct KV cache to maintain accuracy. In this case, it uses a prefetching mechanism, inspired by InfiniGen, to predict KV requirements for the next layer and retrieves the data early. The offloading and fetching process relies on both a zero-copy transfer engine, which is built on GDRCopy developed by Nvidia, and a GPU-centric synchronization method to fully utilize the PCIe bandwidth and eliminate GPU stalls. Similar to other methods, CLO also incorporates outlier handling mechanisms. Certain attention heads have disproportionate influence on model output and are therefore designated as critical. CLO keeps the KV cache of these heads permanently in GPU memory to prevent latency spikes and accuracy degradation. For all other heads, the system follows the similarity‑guided workflow, either reusing previous KV data or fetching new KV entries with prefetching as needed.

\begin{table}[!t]
\centering
\caption{Hybrid Memory Solutions Comparison Table}
\label{tab:hybrid_memory}
\begin{tabularx}{\textwidth}{@{}p{2cm}p{3.5cm}p{5cm}X@{}}
\toprule
\textbf{Method} & \textbf{Offload destination} & \textbf{Mechanism} & \textbf{Key optimization} \\
\midrule
Paged Attention~\cite{PagedAttention} & CPU Memory (DRAM) & Paging the KV cache, inspired by virtual memory & Reducing memory fragmentation and enabling cache sharing \\ 
InfiniGen~\cite{InfiniGen} & CPU Memory (DRAM) & Attention speculation and prefetching & Reducing PCIe transfer volume by only fetching essential data \\ 
LayerKV~\cite{LayerKV} & CPU Memory (DRAM) & Layer‑wise management and SLO‑aware scheduler to balance GPU/CPU residency & Reducing TTFT by minimizing queuing delays \\ 
INF2~\cite{INF2} & Host Memory + NVMe SSDs (CSDs) & Attention‑Near Storage (ANS) via Computational Storage Devices (CSDs) & High throughput for long‑context models by leveraging internal storage bandwidth \\
KVPR~\cite{KVPR} & CPU Memory (DRAM) & Overlapping partial cache recomputation on the GPU with asynchronous transfer & Minimizing GPU idle time caused by limited PCIe bandwidth \\ 
Oneiros~\cite{Oneiros} & CPU Memory (DRAM) & Parameter remapping to repurpose GPU memory from inactive models into KV cache space & Optimizing multi‑tenant serving by reclaiming memory from idle models \\ 
CLO~\cite{CLO} & CPU Memory (DRAM) & Algorithm‑system co‑design with head‑wise approximate caching and zero‑copy transfer engine & Eliminating CPU bottlenecks and fully utilizing PCIe bandwidth \\ 
\bottomrule
\end{tabularx}
\end{table}

\subsection{New Attention Calculation}
The Transformer architecture, introduced in Attention Is All You Need~\cite{Attention}, relies on Scaled Dot-Product Attention as its core mechanism. This method computes a compatibility score between each query and key vector pair to estimate how well they match. It scales the score by the square root of the hidden dimension, and applies a softmax function to obtain normalized attention weights.  These weights are then used to compute a weighted sum of the corresponding value vectors, producing the output representation.  

While this approach has become the foundation of modern LLMs, its computational complexity of $O(n^2)$ with respect to sequence length poses significant challenges for long-context inference. Consequently, recent research has focused on developing alternative attention mechanisms that reduce complexity while preserving accuracy. These innovations include linear attention, log-linear attention, and hybrid methods that balance efficiency and expressiveness. We summarized the attention variants in Table~\ref{tab:attention_variants}.

One foundational line of work seeks to reduce the quadratic cost of softmax attention. Transformers‑are‑RNNs \cite{Linear} is one of the first works that introduces the concept of linear attention for autoregressive transformers, demonstrating how the quadratic $O(N^2)$ complexity of softmax attention can be reduced to linear time O(N). The key idea is to replace the softmax similarity function with a kernel-based linear attention.
For a single query vector q (one token), the typical linear-attention output is written as:
\[
V_i' = \frac{\phi(Q_i)^T \sum_{j=1}^{i} \phi(K_j) V_j^T}
{\phi(Q_i)^T \sum_{j=1}^{i} \phi(K_j)}
\]
It can be simplified into:
\[
V_i' = \frac{\phi(Q_i)^T S_i}{\phi(Q_i)^T Z_i}
\]
where \[
S_i = \sum_{j=1}^{i} \phi(K_j) V_j^T
\] \[
Z_i = \sum_{j=1}^{i} \phi(K_j)
\]

Here, $\phi(\cdot)$ is a feature map. It is a function that is applied to each query or key vector elementwise that produces a new vector of the same size. It can turn the original similarity $(q,k)$ from: similarity $(q,k) = softmax(q^T k)$ into an approximation of the form similarity $(q,k) \approx \phi(q) \cdot \phi(k)$ that can be written as a dot product of transformed vectors.
By expressing similarity as a dot product between transformed vectors, the attention computation can be rearranged into a sequence of associative matrix multiplications. This allows the key–value accumulation term S to be computed incrementally, which effectively transforms the attention mechanism into a recurrence that operates in linear time and memory with respect to sequence length. 

A middle-ground alternative is explored in Log-Linear Attention \cite{LogLinear}, allowing the cost to be between the traditional softmax attention (which has quadratic $O(N^2)$ complexity) and efficient linear-attention variants. Its central idea is to hierarchically summarize past tokens so that attention computation scales in $O(N\log N)$ time and $O(\log N)$ memory.

\begin{figure}[!tb]
    \centering
    \includegraphics[width=0.6\linewidth]{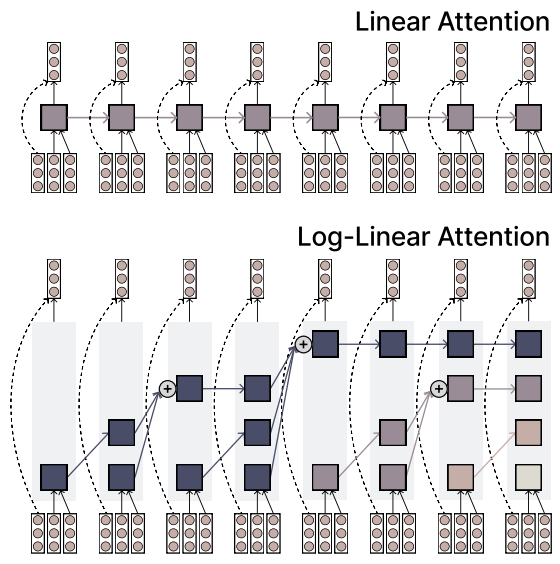}
    \caption{ Standard linear attention (top) vs. loglinear attention (bottom). The input consists of query, key, and value vectors ~\cite{LogLinear}.}
    \label{fig:LogLinear}
\end{figure}

The method organizes token history using a Fenwick-tree–style structure, where past tokens are grouped into buckets whose sizes are powers of two. Each token gets assigned to a bucket level based on its position in the sequence. Each new token is first placed in the finest-grained bucket, while older tokens progressively ``age'' into coarser buckets. For example, when there are 8 tokens, $Bucket_0$ handles the current token, $Bucket_1$ holds the last 1 token, $Bucket_2$ holds the last 2 tokens, and $Bucket_3$ holds the last 4 tokens. Therefore, at time step $t$, there will be $\log_2(t)$ buckets, each representing a summary over a different time scale.
As the sequence grows, buckets automatically merge following the Fenwick-tree update logic, ensuring that at most $\log_2(t)$ buckets are active. This hierarchical nature results in a computational cost of $O(N\log N)$ (log-linear time) and memory cost of $O(\log N)$ (logarithmic space). 
For each bucket $\ell$ at time $t$, the system computes a summary matrix $S_t^{(\ell)}$. Then, each bucket gets a weight and uses it to compute the output. This process is represented in the following formula:
\[
o_t = \sum_{\ell=0}^{L-1} \lambda_t^{(\ell)} q_t^T
\left(\sum_{s \in B_t^{(\ell)}} v_s k_s^T \right)
= \sum_{\ell=0}^{L-1} \lambda_t^{(\ell)} q_t^T S_t^{(\ell)}
\]

A regression‑inspired formulation appears in Local Linear Attention (LLA) \cite{LocalLinear}, where attention is approximated locally, striking a balance between softmax responsiveness and linear‑attention efficiency.  
Local Linear Attention is inspired by the similarity between attention and regression. Conceptually, attention takes past keys and values as “training examples” and produces an output for the current query, similar to how regression predicts a new data point from observations. Under this interpretation, Softmax Attention behaves like local constant regression because it looks at nearby keys and averages their value; while Linear Attention is alike global linear regression because it fits a global straight line for all data. Based on such observation, the authors proposed Local Linear Attention, which is similar to local linear regression. This enables LLA to adapt more effectively to non‑stationary or time-varying patterns in the sequence, offering a middle ground between locally responsive softmax attention and the globally smoothed behavior of linear attention. 
In practice, LLA fits a small local linear model around each query by selecting a neighborhood of relevant key–value pairs and estimating a local linear approximation. The output is then computed based on this locally fitted model, allowing the attention mechanism to respond dynamically to changes in the underlying data distribution while retaining the efficiency benefits of linear‑attention structures. 

A more expressive linear framework is introduced in KIMI Linear \cite{Kimi}, combining dynamic gating mechanisms with a hybrid of linear and softmax layers to improve representational power.

KIMI Linear introduces Kimi Delta Attention (KDA), an expressive form of linear attention designed to improve the representational capacity of linear-time architectures. KDA is the central innovation of the work, and KIMI Linear integrates KDA and full softmax attention layers in a 3:1 hybrid ratio, which the authors identify as the optimal balance between efficiency and model accuracy.

KDA incorporates two dynamic gating mechanisms to control how memory evolves over time. They are forget gate $(\alpha)$ and update rate $(\beta)$. Forget gate $(\alpha)$ is a decay factor that determines how much of the previous memory state $S_{t-1}$ is retained when updating the memory at time step $t$. If $\alpha$ is close to 1, the memory state keeps most of the old info; otherwise, if $\alpha$ is close to 0, it forgets quickly. Update rate $(\beta)$ determines how strongly the new KV pair influences the memory update at time step $t$. Large $\beta$ value means new info dominates, whereas small $\beta$ value means new info has little effect. Both $\alpha$ and $\beta$ are calculated from the input token using small neural networks, whose parameters are learned during model training. In KDA, the memory state is represented by matrix $S$. It is calculated using the following formula:

\[
S_t = (I - \beta_t k_t k_t^T)\,\mathrm{Diag}(\alpha_t)S_{t-1} + \beta_t k_t v_t^T \in \mathbb{R}^{d_k \times d_v}
\]

Where: $S_{t-1}$ is the previous memory state; $\mathrm{Diag}(\alpha_t)$ is a diagonal matrix where each diagonal entry is one element of $\alpha_t$; $(I - \beta_t k_t k_t^T)$ adjusts memory so it doesn’t overfit old associations; $\beta_t k_t v_t^T$ store the new key-value pair in memory. The dynamic memory state allows the model to retain specific information components selectively. And with $S$, the output can be calculated with:

\[
o_t = S_t^T q_t \in \mathbb{R}^{d_v}
\]

Beyond KDA, KIMI Linear incorporates additional optimizations. Instead of updating one token at a time, Kimi Linear processes tokens in chunks to increase computational parallelism and improve throughput. Also, KIMI Linear combines KDA (linear attention) and full attention at 3:1 ratio, which the authors empirically find delivers the best trade-off between efficiency and performance, preserving accuracy while significantly reducing memory and compute cost.

\begin{table}[!tb]
\centering
\caption{Attention Variants -- Mechanisms, Complexities, and Features}
\label{tab:attention_variants}
\begin{tabularx}{\textwidth}{@{}p{2cm}p{4cm}p{1.5cm}p{1.5cm}p{1.5cm}X@{}}
\toprule
\textbf{Method} & \textbf{Mechanism} & \textbf{Training Time Complexity} &
\textbf{Decoding Time Complexity per step} & \textbf{Decoding Space Complexity (Total Memory)} &
\textbf{Features} \\
\midrule
Softmax &
Scaled Dot‑Product Attention \& Multi‑Head Attention &
$O(T^2)$ &
$O(T)$ &
$O(T)$ &
High expressivity, high cost \\
Linear &
Replaces softmax with a linear dot‑product of kernel feature maps &
$O(T)$ &
$O(1)$ &
$O(1)$ &
Limited expressivity and low cost \\
Log Linear &
Replaces the fixed‑size hidden state with a logarithmically growing set of hidden states &
$O(T \log T)$ &
$O(\log T)$ &
$O(\log T)$ &
Balances efficiency and expressiveness \\
Local Linear &
Use Local Linear Regression to perform query‑specific local linear fitting &
$O(T^2)$ &
$\sim O(T)$ &
$O(T)$ &
Superior bias–variance trade‑off; high cost \\ 
KIMI Linear &
Hybrid architecture (KDA + Multi‑Head Latent Attention) &
Mainly $O(T)$ due to KDA &
$O(1)$ &
$O(1)$ &
Outperforms full attention, but requires hybridization (e.g., 3:1 ratio with full attention) to maintain global information flow \\ 
\bottomrule
\end{tabularx}
\end{table}

\subsection{Combination Methods}
Combination methods integrate multiple optimization strategies to achieve greater efficiency in KV cache management than any single technique alone. These approaches often combine eviction, compression, and hybrid memory solutions to balance memory savings, computational cost, and accuracy. For example, some methods apply token selection algorithms alongside quantization to reduce cache size, whereas others combine compression with offloading to CPU or disk to handle extremely long contexts. By leveraging complementary techniques, combination strategies aim to overcome the limitations of individual methods and deliver scalable solutions for large-context inference. 

FlexGen\cite{FlexGen} combines compression and offloading for extreme resource constraints. It enables inference for massive LLMs using only one single GPU with limited memory. This is done by splitting model weights, activations, and KV cache across the GPU, CPU memory, and disk. To determine the optimal placement and movement of these components, FlexGen constructs a cost model that estimates latency, bandwidth, and memory constraints for each device. It then solves a linear programming optimization problem to minimize the per‑token generation time, yielding an execution plan that balances compute and I/O overhead. 
To further reduce memory footprint, FlexGen applies 4‑bit compression to both model weights and the KV cache using group‑wise quantization, where each small group of values shares a common scaling range (min/max). This achieves substantial compression with minimal accuracy degradation.  
It needs to be noted that FlexGen is designed specifically for high‑throughput workloads, such as batched prompt processing, rather than low‑latency, single‑request scenarios. 
To sustain large batch sizes on limited GPU memory, it uses a zig‑zag block scheduling strategy, which means processing multiple batches across one computational layer before moving to the next layer, instead of processing an entire batch through the full model sequentially. This strategy maximizes the reuse of model weights already loaded onto the fast GPU memory, thus reducing repeated loading and minimizing slow I/O transfers. 

A sparsity‑aware quantized design is presented in Q‑Hitter \cite{Q-Hitter}, where tokens are selected based on both attention importance and quantization robustness. In other words, the token‑selection mechanism identifies tokens that are both semantically important and amenable to quantization, enabling efficient storage of a sparse, quantized KV cache. For each new token, the system computes two metrics based on its key and value vectors: the attention score and quantization error. 
Attention score reflects the token’s importance for future predictions by measuring how strongly it is likely to influence subsequent queries. Quantization error shows the amount of information lost if the token’s KV pair were stored in a low‑bit representation. 
These two measurements are combined, via a balancing mechanism introduced in the paper, into a unified score S. Tokens with high S values are both important and robust to quantization noise. Therefore, Q‑Hitter selects the top‑K tokens according to S and stores only their KV entries in quantized form. Tokens with lower scores are discarded, resulting in a sparse KV cache that preserves critical information while significantly reducing memory footprint and I/O cost. 

\begin{figure}[!tb]
    \centering
    \includegraphics[width=1\linewidth]{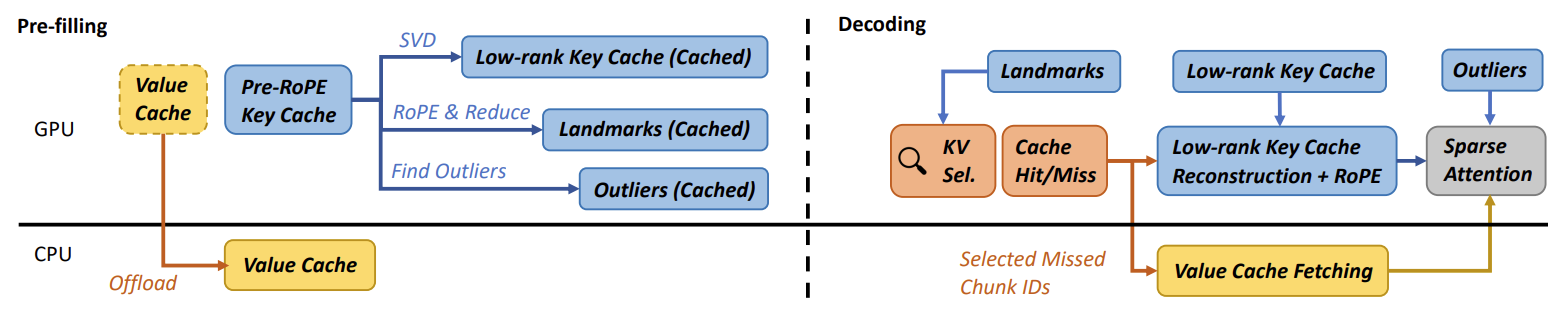}
    \caption{ During Pre-filling, ShadowKV offloads the value cache to the CPU while maintaining a low-rank key cache, landmarks, and outliers on the GPU. During decoding, it employs landmarks for sparse attention. ~\cite{ShadowKV}.}
    \label{fig:ShadowKV}
\end{figure}

\begin{figure}[!tb]
    \centering
    \includegraphics[width=1\linewidth]{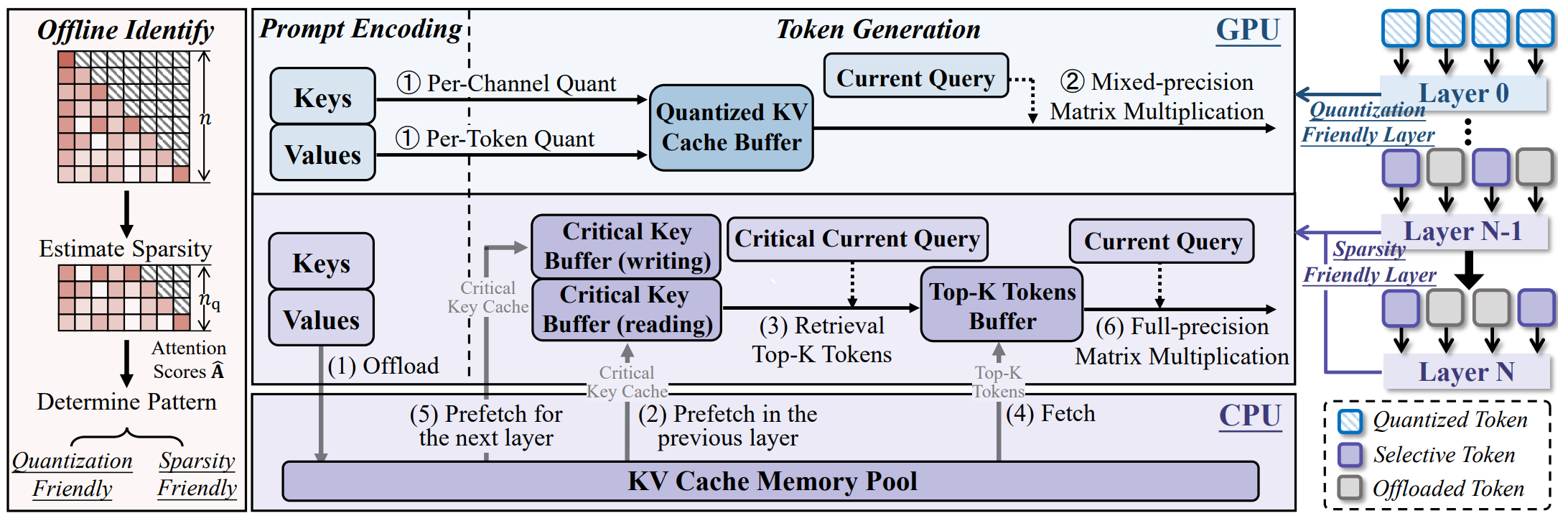}
    \caption{System overview of TailorKV. Offline identification categorizes the layers into quantization-friendly and sparsity-friendly. For quantization-friendly layers, we employ aggressive static quantization. For sparsity-friendly layers, we dynamically retrieve Top-K tokens. Critical current query and critical key cache represent the outliers in the query and key cache, respectively. ~\cite{TailorKV}.}
    \label{fig:TailorKV}
\end{figure}

ShadowKV \cite{ShadowKV} is a hybrid approach that adopts both cache compression and hybrid memory It stores compressed keys on GPU while offloads values to CPU memory. The key insight is that pre‑RoPE keys exhibit strong low‑rank structure, enabling effective compression without accuracy degradation.  
During the prefill phase, ShadowKV applies Singular Value Decomposition (SVD) to the pre‑RoPE key matrix, retains only the top‑rank components, and stores these compressed keys in GPU memory.  
In contrast, value vectors do not exhibit low‑rank properties and therefore cannot be compressed effectively. To reduce GPU memory usage, ShadowKV offloads values to CPU memory.  
To enable efficient selective retrieval during decoding, the system divides post‑RoPE keys into chunks and computes a landmark vector (typically the mean) for each chunk. It then compares each token’s post‑RoPE key with its corresponding landmark. The tokens, whose keys showing low cosine similarity with the landmark, are marked as outliers because they cannot be well-represented by the chunk summary. These outlier tokens are fully stored on the GPU.  During the decoding phase, ShadowKV uses the landmark vectors to estimate which chunks are likely to contribute most to the current attention computation. Only the values corresponding to these important chunks are fetched from CPU memory. Meanwhile, the compressed pre‑RoPE keys stored on GPU are decompressed to reconstruct the necessary key vectors. Once both the reconstructed keys and the fetched values are available, the system performs the standard attention computation to generate the next token.

Layer‑specific strategies are combined in TailorKV\cite{TailorKV}. TailorKV as shown in Fig.~\ref{fig:TailorKV} is a hybrid KV‑cache optimization framework that exploits a key empirical observation about transformer layers. That is, in transformer models, shallow layers (closer to the input) tend to distribute their attention broadly across the sequence and are therefore more amenable to quantization; while deeper layers focus attention on a small subset of critical tokens, making them better suited for sparsity‑based offloading. TailorKV combines these complementary strategies to support efficient long‑context inference. During the prefill phase, TailorKV analyzes each layer’s attention distribution. It computes attention scores betwwen the last query against all keys in that layer. Then it selects the top-k attention scores and sums them. The sum of these top-k values reflects how concentrated the layer’s attention is. A small sum indicates that this layer’s attention is spread out, suggesting that the layer is quantization‑friendly. A large sum indicates that the attention is concentrated on a few tokens, meaning the layer is sparsity‑friendly. Based on this metric, TailorKV assigns each layer to one of two regimes.  For quantization-friendly layers, the KV cache will be compressed aggressively and stored compactly on GPU, reducing memory footprint while preserving accuracy due to their inherent global focus. 
For the sparsity-friendly layers, their KV caches are offloaded to CPU memory during prefill phase. Among them, only the most critical tokens (identified dynamically) will be fetched to GPU during decoding.  
To ensure high throughput, TailorKV uses double buffering, overlapping CPU-GPU data transfers with GPU computation, so that fetching and attention updates proceed concurrently.

\section{Comparative Analysis} \label{sec:comparative_analysis}

{\small
\begin{longtable}{p{2.5cm}p{2.5cm}p{3.5cm}p{2.5cm}p{3.6cm}}
\caption{Comparison of KV Cache Optimization Techniques}
\label{tab:kv_cache_optim}\\
\toprule
\textbf{Technique} & \textbf{Memory} & \textbf{Speedups} & \textbf{Accuracy loss} & \textbf{Tradeoffs} \\
\midrule
\endfirsthead

\toprule
\textbf{Technique} & \textbf{Memory} & \textbf{Speedups} & \textbf{Accuracy loss} & \textbf{Tradeoffs} \\ 
\midrule
\endhead

\midrule
\multicolumn{5}{r}{\small\itshape Continued on next page} \\
\endfoot

\hline
\endlastfoot

H$_2$O~\cite{H2O} &
Up to 5--10$\times$ memory reduction &
Up to 29$\times$ throughput improvement; $\le$ 1.9$\times$ lower latency compare to FlexGen &
Comparable to baseline &
accumulated-attention bias; risk of heavy‑hitter loss \\ 

SnapKV~\cite{SnapKV} &
8.2$\times$ memory efficiency &
3.6$\times$ generation speedup &
Comparable to baseline &
Does not optimize prefill; cannot extend inherent model context limits \\ 

NACL~\cite{NACL} &
Up to 5$\times$ KV reduction &
O(1) single-shot eviction; major prefill simplification &
95\% performance retention &
Proxy-token selection heuristic; limited ultra-long behavior \\ 

InfiniPot~\cite{InfiniPot} &
Tested with 4k to 1M tokens &
“Performance regardless of context length” &
Consistent performance regardless of context length and optimizing memory use &
Fixed compression ratio may not be optimal for all data types \\ 

HASHEVICT~\cite{HashEvict} &
30--70\% compression &
1.5--2$\times$ prefill speedup against baseline methods such as H$_2$O, Scissorshands, and 17$\times$ prefill/2$\times$ decoding speed against FastGen &
Decent performance except at very small budgets (e.g., 10\% capacity) &
Irreversible eviction; quality drops sharply at 10\% cache budget \\ 

MorphKV~\cite{MorphKV} &
Up to 5$\times$ over Full-Attention. &
Up to 4.68$\times$ faster than SnapKV &
Comparable to SnapKV &
Performance is sensitive to hyperparameters like window size and fusion functions \\ 

RocketKV~\cite{RocketKV} &
Up to 400$\times$ compression &
Up to 3.7$\times$ speedup &
Negligible accuracy loss &
Focused on decode; limited prefill optimization \\ 

KVzip~\cite{KVzip} &
Up to 70\% eviction &
2$\times$ reduction in FlashAttention decoding latency &
Negligible accuracy loss &
Reconstruction overhead (amortizable over multiple queries); no formal guarantees on information loss \\ 

Ada-KV~\cite{Ada-KV} &
4$\times$ cache reduction &
Comparable decoding latency to SnapKV &
Slight quality improvement ($\sim$5\%) &
Allocation only within layers, not across model \\ 

KIVI~\cite{KIVI} &
2.6$\times$ peak memory reduction &
2.35$\times$--3.47$\times$ throughput &
$<$2\% accuracy drop for most models &
Models with one KV head may require 4‑bit quantization to maintain accuracy; quantization overhead due to initial cost for the quantization \\ 

MiniCache~\cite{MiniCache} &
Up to 41\% memory reduction &
$\sim$5$\times$ throughput &
Minimal loss &
Limited to merging only two layers at a time; restricting higher compression \\ 

PALU~\cite{PALU} &
$\sim$50\% KV compression &
Up to 1.89$\times$ (RoPE) or 2.91$\times$ (with quantization) &
Comparable to baseline &
Reconstruction overhead for keys (especially with RoPE) \\ 

KVQuant~\cite{KVQuant} &
3.7--6.9$\times$ memory savings &
Up to $\sim$1.7$\times$ speedup &
$<$0.1 perplexity degradation with 3‑bit quantization &
Long‑context training challenges; more complex dequantization process \\ 

PagedAttention~\cite{PagedAttention} &
Offload based (no reduction) &
2--4$\times$ higher throughput with the same latency &
Lossless &
Kernel overhead; computational overhead for managing logical to physical mapping \\ 

InfiniGen~\cite{InfiniGen} &
Offload based (no reduction) &
1.63--32.9$\times$ speedup &
Comparable accuracy with $>$15\% Relative KV Cache Size &
Requires additional GPU memory to store partial weights used for speculation; Specifically designed for CPU-offloading systems \\ 

LayerKV~\cite{LayerKV} &
Offload based (no reduction) &
Up to 69$\times$ TtFT improvement &
Lossless &
Slight throughput trade-off under high-load decoding \\ 

INF2~\cite{INF2} &
Offload based (no reduction) &
3.46$\times$ throughput; KV I/O overhead reduced by $>$80\% &
Lossless &
CPU coordination overhead; Requires CSD hardware \\ 

KVPR~\cite{KVPR} &
Offload based (no reduction) &
Up to 35.8\% lower latency, 46.2\% higher throughput compared to DeepSpeed and HuggingFace Accelerate &
Lossless &
Decoding focus only; limited to single-GPU or data-parallel setups \\ 

Oneiros~\cite{Oneiros} &
Offload based (no reduction) &
44.8\%--82.5\% time-between-token latency reduction; 20.7\%--99.3\% TtFT reduction; 6.6\%--86.7\% throughput improvement compared to vLLM &
Lossless &
Requires high bandwidth for CPU--GPU \\ 

CLO~\cite{CLO} &
Offload based (no reduction) &
9.3\%--66.6\% throughput improvement compared to SOTA baseline (RetroInfer and InfiniGen) &
Near lossless: $\le$0.42 accuracy drop compared to original models &
Approximation leads to sacrificing a small amount of cache hit ratio to gain speed; Hyperparameter tuning required manual tuning; PCIe 4.0 dependency \\ 

LinearAttention~\cite{Linear} &
O(1) memory &
Up to 4000$\times$ faster on long sequences compared to vanilla transformers &
Strong on synthetic tasks; weaker on complex tasks &
Performance sensitive to choice of feature map kernels; accuracy gap for reasoning \\ 

Log Linear Attention~\cite{LogLinear} &
$O(\log T)$ memory &
3$\times$ speedup over naive implementations &
Better than linear; below full attention &
Performance gap compared to full attention; engineering complexity \\ 

Local Linear Attention~\cite{LocalLinear} &
Similar to softmax &
Faster than softmax &
Outperforms Softmax Attention and Linear Attention on associative and regression tasks &
Trades off speedups for accuracy; limited memory benefit \\ 

KIMI~\cite{Kimi} &
Up to 75\% KV reduction &
Up to 6$\times$ throughput at 1M context &
Outperforms full attention &
Kernel dependency for maximum efficiency \\ 

FlexGen~\cite{FlexGen} &
Up to 10$\times$ memory reduction &
40$\times$ to 100$\times$ higher maximum throughput compared to DeepSpeed Zero-Inference and HuggingFace Accelerate &
Negligible accuracy loss with 4-bit &
High latency; PCIe/Disk bandwidth bottleneck for KV retrieval; insufficient for small batches \\ 

Q-Hitter~\cite{Q-Hitter} &
Up to 20$\times$ memory reduction &
Up to 33$\times$ vs HF Accelerate &
Full quality preservation &
Computational overhead for calculating quantization errors and performing dequantization \\ 

ShadowKV~\cite{ShadowKV} &
6$\times$ GPU memory reduction &
3.04$\times$ throughput improvement &
High accuracy until sparse budget $<$1.56\% &
Performance partially dependent on PCIe bandwidth \\ 

TailorKV~\cite{TailorKV} &
$\sim$73.8\% GPU memory reduction &
8--18$\times$ faster than standard offloading &
Near lossless &
Prefill bottleneck; system complexity \\ 
\bottomrule

\end{longtable}
}

\section{Scenarios} \label{sec:scenarios}
\subsection{Long Context (>1M tokens) for Single Request:}
Cache eviction and cache compression are generally well-suited for handling long-context workloads involving single requests. In such settings, the primary bottleneck arises from the substantial memory footprint associated with processing extended sequences. Because cache eviction and cache compression are explicitly designed to substantially reduce the KV cache memory requirements, while incurring only negligible accuracy degradation, they represent the most appropriate optimization strategies. Hybrid techniques that incorporate these two methods are likewise suitable for this scenario. 

Beyond these categories, KIMI from category 4 is also applicable, as its authors explicitly note that Kimi Linear is designed to process million-token sequences while achieving up to 6.3× higher throughput and reducing memory usage by 75\% compared to standard models. 

\subsection{Minimal model modification:}
Cache eviction and cache compression are generally recommended for minimal model modification, as both approaches directly target reductions in memory footprint. Among these methods, Ada-KV, SnapKV, and KIVI are particularly suitable. These techniques are fine-tuning-free and are designed for seamless “plug-and-play” integration into existing models with minimal architectural adjustments. If compression or quantization is preferred over eviction, KIVI represents the most appropriate choice, as it offers a tuning-free 2-bit solution capable of handling outliers without necessitating model retraining. 

Linear, log-linear, local linear, and KIMI methods are generally not recommended for scenarios requiring minimal model modification. These methods involve fundamental architectural changes, new attention mechanisms, or extensive training, which contrast with the ``plug-and-play'' or ``training-free'' nature of post-training KV cache management strategies. 

\subsection{High throughput serving:  }
Hybrid memory solutions are generally well-suited for achieving high throughput. Among the methods in this category, PagedAttention (vLLM) is the most appropriate, as it is specifically designed to maximize throughput by enabling a larger number of requests to be processed within a single batch. Similarly, Oneiros targets multi-tenant LLM serving and attains up to 86.7\% higher throughput than vLLM by remapping parameter memory for KV‑cache usage. 

ShadowKV is also a strong candidate, as it integrates cache compression with hybrid memory techniques. By offloading values and employing low-rank keys, ShadowKV supports batch sizes up to 6× larger and improves throughput by up to 3.04×. 

For scenarios requiring extreme throughput in resource‑constrained environments (e.g., a single GPU), FlexGen and its successor Q-Hitter utilize aggressive offloading strategies. These systems achieve up to 33× higher throughput than standard offloading frameworks through the co-design of quantization and sparsity.

\subsection{Edge/memory-limited devices }
For edge or memory‑limited devices, cache eviction and cache compression are also effective techniques, as they reduce memory pressure by minimizing the overall memory footprint. Among the methods in this category, InfiniPot is explicitly designed for on‑device environments (mobile/edge) where VRAM or NPU memory constitutes a strict constraint; it employs Continual Context Distillation to maintain the KV cache within a fixed‑size “pot.” TailorKV likewise targets resource‑limited GPUs and was specifically developed to serve 128k‑context 8B models on a single RTX 3090 GPU. 

Conversely, methods with demanding resource requirements are not preferred due to the inherently constrained nature of edge environments. Most hybrid‑memory approaches fall into this scenario. For example, PagedAttention is optimized for high‑end GPUs, while Oneiros depends heavily on extremely high CPU–GPU bandwidth (450–900 GB/s), which is rarely available on edge devices. These methods are therefore unsuitable for edge deployment due to their hardware requirements.

\subsection{Multi-turn Conversations:  }
Standard cache‑eviction methods such as H$_2$O are not suitable for multi‑turn conversations, as they permanently discard tokens that may be required in later turns. However, several methods are specifically optimized for such dialogue-oriented scenarios. For example, RocketKV‑MT is a variant of the standard RocketKV designed for multi‑turn settings; it retains all KV tokens in memory for future turns while constraining token selection in the current turn. KVzip is a query‑agnostic method that optimizes reusable compressed KV caches, enabling efficient inference across diverse future queries within a dialogue. The authors note that its “overhead can be amortized over multiple queries.” ShadowKV is another strong option, as it demonstrates “multi‑turn capability” by sharing low‑rank subspaces between a sequence and its continuation, thereby maintaining accuracy across multiple rounds of interaction. 

Besides standard cache eviction, methods that impose substantial latency are also poorly suited for this setting, as multi‑turn conversations also require low latency and responses must be generated quickly to maintain an interactive user experience. For example, FlexGen is optimized for latency‑insensitive batch processing, and its inherent delays render it impractical for interactive dialogue. 

\subsection{Prefill-Heavy Workloads:  }
NACL is recommended because it performs efficient eviction in a single operation during the encoding phase, rather than step-by-step during decoding. This design reduces time complexity and minimizes the computational overhead typically associated with managing large‑scale KV caches. 

HASHEVICT is also a strong candidate, as it achieves a 1.5×–2× prefill speedup by making eviction decisions prior to the attention computation. 

If the primary bottleneck is Time to First Token (TTFT) for very long prompts, LayerKV (69× improvement in TTFT) or MiniCache (significant prefill efficiency) are recommended. In the LayerKV paper, the authors state that ``Comprehensive evaluations on representative models, ranging from 7B to 70B parameters, across various GPU configurations, demonstrate that LayerKV improves TTFT latency up to 69$\times$.'' \cite{LayerKV} 

Lastly, CLO processes the full prompt in parallel during the prefill phase and employs speculative sparse prefetching to hide subsequent loading overhead, making it a suitable option as well. 

\subsection{Accuracy-Critical Reasoning:  }
Hybrid memory solutions such as PagedAttention are preferred for accuracy‑critical reasoning tasks. Methods in this category improve overall efficiency by leveraging hybrid memory infrastructures rather than modifying data representations or altering model components. As a result, they maintain lossless accuracy, making them suitable for scenarios where reasoning precision is essential. 

Conversely, cache‑eviction and compression‑based techniques reduce memory footprint at the cost of accuracy. Likewise, linear attention and log‑linear attention methods tend to underperform in highly complex reasoning scenarios when compared to full attention. Therefore, these approaches should be avoided in accuracy‑critical applications. 

\subsection{Hardware-Specific Constraints:  }
For configurations with high PCIe bandwidth (e.g., NVIDIA GH200 systems), Oneiros and CLO are superior choices because they leverage the interconnect to offload KV caches with negligible overhead. Oneiros exploits the high CPU–GPU bandwidth available on specialized hardware such as the NVIDIA Grace Hopper (GH200) to repurpose parameter memory for KV‑cache storage, while CLO utilizes zero‑copy transfer engines and custom CUDA kernels to fully exploit PCIe bandwidth on modern GPU platforms. Therefore, both methods are suitable for this scenario. 

INF2 is designed for fast inference using Computational SSDs (CSDs) equipped with ASICs or FPGAs to offload KV‑related operations. When such hardware is available, this method should be considered the top choice. 

FlexGen is presented as a high‑throughput generation engine capable of running LLMs with limited GPU memory. It can be flexibly configured under various hardware resource constraints by aggregating memory and computation across the GPU, CPU, and disk.

\section{Summary} \label{sec:summary}

We categorized KV‑cache optimization techniques into the following categories: cache eviction, cache compression, hybrid memory solutions, new attention mechanisms, and combination methods. This survey also reveals distinct strengths, limitations, and ideal deployment scenarios for each.  

First, we observe that eviction, compression, and their combination approaches increasingly dominate ultra‑long‑context scenarios, particularly when context lengths exceed one million tokens. Methods such as RocketKV, KVzip, and ShadowKV achieve high compression ratios with minimal accuracy degradation by combining coarse‑grained token selection with fine‑grained reconstruction or low‑rank approximations. These methods substantially reduce GPU memory pressure while retaining global contextual fidelity, making them well‑suited for large‑context, single‑request workloads.  

Second, hybrid memory solutions are essential for scaling beyond physical GPU limits through CPU offloading and efficient block-based memory allocation because they fundamentally change where KV resides rather than what is stored. This category of methods consistently excels under high‑throughput or multi‑tenant serving conditions. Systems like PagedAttention, Oneiros, KVPR, and INF2 demonstrate that offloading, remapping, and near‑storage computation are more effective than data‑centric optimizations when throughput is the primary metric. Their performance benefits are largely orthogonal to model structure, enabling lossless accuracy and scalable serving across diverse workloads. Therefore, hybrid methods are the strongest fit for data‑center deployments requiring high concurrency, where accuracy remains lossless, and hardware interconnect speed can be fully exploited. 

Alternative attention mechanisms, including linear, log‑linear, local linear, and KIMI, offer asymptotically better scaling than standard softmax attention, reducing complexity from $O(N²)$ to $O(N \log N)$ or $O(N)$. These methods are promising for architectural redesigns where ultra‑long context is a first‑class goal. However, they require full model retraining and still lag in accuracy‑critical reasoning, making them less suitable as drop‑in optimizations for existing LLMs. Their long‑term significance lies in pointing to future transformer successors that natively support million‑token sequences without relying on cache pruning. 

Hybrid strategies such as FlexGen, Q‑Hitter, ShadowKV, and TailorKV demonstrate that no single optimization dominates across all scenarios. By mixing sparsity, quantization, compression, and offloading, these frameworks deliver efficiency unattainable by individual techniques, particularly under extreme resource constraints. Combination methods are especially effective when workloads demand balanced latency, throughput, and memory savings, or when operating on consumer‑grade GPUs with limited VRAM. 

Overall, across the surveyed landscape of KV‑cache optimization techniques, several clear patterns emerge. Eviction and compression strategies dominate for ultra‑long‑context workloads, while hybrid memory systems dominate for high‑throughput serving. Standalone compression excels in bandwidth‑bound environments, and new attention mechanisms provide a path toward fundamentally scalable architectures. Ultimately, the future of long‑context inference lies in integrated, multi‑stage KV optimization pipelines that adapt dynamically to context length, system load, and hardware constraints.


\bibliographystyle{IEEEtran}
\bibliography{references}

\end{document}